%% file: main.tex
\documentclass[twocolumn]{article}

\usepackage{PRIMEarxiv}

\usepackage[utf8]{inputenc} 
\usepackage[T1]{fontenc}    
\usepackage{hyperref}       
\usepackage{url}            
\usepackage{booktabs}       
\usepackage{amsfonts}       
\usepackage{nicefrac}       
\usepackage{microtype}      
\usepackage{amsmath}
\usepackage{amssymb}
\usepackage{lipsum}
\usepackage{fancyhdr}       
\usepackage{graphicx}       
\graphicspath{{media/}}     
\usepackage{setspace}
\usepackage{authblk}
\usepackage{csquotes}
\usepackage{siunitx}
\usepackage{todonotes}
\usepackage{subcaption}
\usepackage{acronym}        
\DeclareSIUnit\pixel{px}
\newlength{\plotwidth}
\newcommand\hide[1]{}
\newboolean{show_notes}
\setboolean{show_notes}{true}
\newcommand\note[1]{\ifthenelse{\boolean{show_notes}}{\textcolor{red}{\textbf{Note: }#1}}{\hide{#1}}}

\newcommand{\mP}{\mathbb{P}}

\newcommand{\bx}{\boldsymbol{x}}
\newcommand{\KLD}{\operatorname{\ac{KLD}}}
\newcommand\sugarvit{SugarViT}

\title{\sugarvit{}\textemdash Multi-objective Regression of UAV Images with Vision Transformers and Deep Label Distribution Learning Demonstrated on Disease Severity Prediction in Sugar Beet}

\author[*,1,2]{\textbf{Maurice Günder}}
\author[3]{\textbf{Facundo R. Ispizua Yamati}}
\author[3]{\textbf{Abel A. Barreto Alcántara}}
\author[3]{\textbf{Anne-Katrin Mahlein}}
\author[1]{\textbf{Rafet Sifa}}
\author[1,2]{\textbf{Christian Bauckhage}}

\affil[1]{Fraunhofer Institute for Intelligent Analysis and Information Systems IAIS, Schloss Birlinghoven, 53757 Sankt Augustin, Germany}
\affil[2]{Institute for Computer Science III, University of Bonn, Friedrich-Hirzebruch-Allee 5, 53115 Bonn, Germany}
\affil[3]{Institute for Sugar Beet Research (IfZ), Holtenser Landstraße 77, 37079 Göttingen, Germany}

\affil[*]{Corresponding author, e-Mail: \url{mguender@uni-bonn.de}}

\begin{document}

\twocolumn[
  \begin{@twocolumnfalse}
    \maketitle
    \begin{abstract}
        Remote sensing and artificial intelligence are pivotal technologies of precision agriculture nowadays. The efficient retrieval of large-scale field imagery combined with machine learning techniques shows success in various tasks like phenotyping, weeding, cropping, and disease control. This work will introduce a machine learning framework for automatized large-scale plant-specific trait annotation for the use case disease severity scoring for \ac{CLS} in sugar beet. With concepts of \ac{DLDL}, special loss functions, and a tailored model architecture, we develop an efficient Vision Transformer based model for disease severity scoring called \sugarvit{}. One novelty in this work is the combination of remote sensing data with environmental parameters of the experimental sites for disease severity prediction. Although the model is evaluated on this special use case, it is held as generic as possible to also be applicable to various image-based classification and regression tasks. With our framework, it is even possible to learn models on multi-objective problems as we show by a pretraining on environmental metadata.
    \end{abstract}
    \vspace{24pt}
    \end{@twocolumnfalse}
    ]


\section{Introduction}\label{sec:background_motivation}

In precision agriculture, the use of \acp{UAV} equipped with multispectral cameras for monitoring agricultural fields is well-established for various tasks regarding plant phenotyping and health status~\cite{abel_disease_incidence, multispectral_phenotyping_1, multispectral_phenotyping_2}. Especially in phenotyping for breeding, one of the main advantages of \ac{UAV} imagery is the mapping flexibility in comparison to satellite imagery, automation and homogenization of laborious and time-consuming visual scoring activities, which usually require large numbers of hours of specialized human labor to score large fields. In agriculture, the term \enquote{visual scoring} is commonly used to describe a field assessment, such as phenotyping canopy structure or the quantification of disease intensity, specifically, \ac{DS}~\cite{phytopathometry}. On the other hand, in data science, a related procedure called \enquote{annotation} is used. Annotation consists of labeling data elements in order to add semantic information or metadata. In essence, although the terms differ in their apparent applications, they represent in this paper an equivalent concept, and in this context, we will use \enquote{annotation} as a synonym for \enquote{scoring}. Theoretically, when large field experiments are conducted, this data is immediately available. However, a lot of human-powered effort is needed to gain information out of large imagery. This is where machine learning comes into play. (Rehashed) \ac{UAV} image data has the potential to serve as training data for even large image-processing deep learning models as the currently broadly applied transformer-based models~\cite{transformers_survey}. The origin of the transformer architecture lies in the field of language processing and has led to large success in recent large language models such as the \ac{GPT}~\cite{transformers_in_nlp}. The basic principle of transformers is the so-called attention mechanism~\cite{attention}. It enables the model to connect and associate features over large semantic or sequential distances. This is beneficial not only for one-dimensional tasks as language processing, since this context can be transferred to higher dimensional use cases like image processing. In this case, we are dealing with a \ac{ViT}~\cite{vit} model.

With the power of transformers also a major drawback appears, namely their low data efficiency. Transformers need lots of data to train. This is why their success is currently mainly in application fields where large datasets are available, such as text data. However, we will show, that also on large-scale agricultural datasets enabled by \acp{UAV}, those models can be used for annotation tasks. To demonstrate this potential in our work, we will focus on a classification task based on single plant images extracted from recorded sugar beet fields according to Günder et al.~\cite{plantcataloging}. The single sugar beet plant images are annotated with \ac{DS} estimations of \acf{CLS}, a fungal leaf disease that is a relevant disease causing yield losses in sugar beet production~\cite{cercospora}. We aim a \ac{DS} prediction modeling task and will motivate a multi-objective approach as well as the use of a deep learning architecture based on a \ac{ViT}. By identifying the \ac{DS} prediction as an ordinal classification, we reinterpret the classification as a regression task by using the concept of \acf{DLDL} introduced in~\cite{dldl}. We further optimize the vanilla \ac{DLDL} approach by an improved loss function that does not need a proper hyperparameter tuning~\cite{full_kl_loss}. After the training, our model, we further call \sugarvit{}, is able to predict the disease severity of individual plant images by a probability distribution which gains training robustness and output interpretability. In the following work, we will go into all the details of \sugarvit{} and the conducted experiments.

\section{Materials and Methods}

\subsection{Data and Preprocessing}

A major challenge that comes with the application and, particularly, the training of vision transformer based architectures requires large amounts of image data. In principle, the utilization of \ac{UAV} imaging from crop fields has the potential to gain those large datasets. However, the conditions under which the images are taken can be very diverse, e.g., due to variable weather, lighting, and resolution. Additionally, device-specific properties can come into play when dealing with different camera models or calibration methods. In the context of plant phenotyping, it is particularly desirable to accumulate image data from multiple growing seasons, which implies that all the above-mentioned difficulties can play a role for an accumulation of large-scale datasets. Thus, in order to utilize as much potential from the data, a preprocessing is needed, that is robust against as many confounding factors as possible.

\subsubsection{Available Field Data}\label{sec:field_data}

The dataset we use in this work consists of multispectral images expressed in reflectance of single sugar beet plants recorded by \ac{UAV} systems on 6 different locations near Göttingen, Germany (\ang{51;33;}N \ang{9;53;}E)~\cite{plantcataloging}. \ac{UAV} systems are equipped with a multispectral camera recording 5 spectral channels. Those are, sorted by wavelength, \textit{blue}, \textit{green}, \textit{red}, \textit{red edge}, and \textit{near infrared}. Due to a large amount of different sensors, sugar beet varieties, resolutions (or ground sampling distances), locations, and time points, the dataset is very diverse. In total, it covers 4 harvesting periods from 2019 to 2022 and comprises 17 different experiments or flight missions. Table~\ref{tab:data_overview} gives an overview over all important information of the dataset. Additionally, Table~\ref{tab:gsd_bands} shows the spectral bandwidths of the two camera sensor systems used in this work.

\begin{table*}[t]
    \small
    \begin{center}
        \begin{tabular*}{\textwidth}{l@{\extracolsep{\fill}}l@{\extracolsep{\fill}}l@{\extracolsep{\fill}}l@{\extracolsep{\fill}}S[table-format=2.1]@{\extracolsep{\fill}}S[table-format=2.0]@{\extracolsep{\fill}}S[table-format=3.0]@{\extracolsep{\fill}}S[table-format=6.0]@{\extracolsep{\fill}}l@{\extracolsep{\fill}}}\toprule
            \textbf{ID} & \textbf{sowing date} & \textbf{location}    & \textbf{sensor}     & \textbf{ground sampling dist.} & \textbf{\# varieties} & \textbf{\# rec. dates} & \textbf{\# images} & \textbf{used for} \\
                        &                      &                      &                     & \si{\milli\meter\per\pixel}    &                       &                        &                    & \\\midrule
                Tr01 & 2019-04-09 & Holtensen I     & RedEdge   & 3.5   & 2   & 23 & 12952  & train/val \\ 
                Tr02 & 2019-04-09 & Holtensen I     & RedEdge   & 15    & 2   & 21 & 19593  & train/val \\ 
                Tr03 & 2020-04-06 & Weende          & ALTUM     & 3     & 2   & 25 & 34143  & train/val \\ 
                Tr04 & 2020-04-06 & Weende          & ALTUM     & 4     & 1   & 25 & 250224 & train/val \\ 
                Tr05 & 2021-04-01 & Sieboldshausen  & ALTUM     & 5.1   & 51  &  7 & 39475  & train/val \\ 
                Tr06 & 2021-04-23 & Holtensen I     & ALTUM     & 3     & 5   & 27 & 89082  & train/val \\ 
                Tr07 & 2021-04-23 & Holtensen I     & ALTUM     & 4     & 1   & 23 & 216098 & train/val \\ 
                Tr08 & 2021-04-23 & Holtensen I     & ALTUM     & 3     & 1   & 22 & 39118  & train/val \\ 
                Tr09 & 2021-04-23 & Holtensen I     & ALTUM     & 5     & 1   & 22 & 48915  & train/val \\ 
                Tr10 & 2021-04-30 & Dransfeld       & ALTUM     & 4     & 1   & 22 & 32604  & train/val \\ 
                Tr11 & 2021-04-30 & Dransfeld       & ALTUM     & 5     & 1   & 22 & 30530  & train/val \\ 
                Te01 & 2022-04-19 & Weende          & ALTUM     & 4     & 1   & 12 & 128550 & test      \\ 
                Tr12 & 2022-04-20 & Reinshof        & ALTUM     & 5     & 4   & 11 & 21036  & train/val \\ 
                Tr13 & 2022-04-20 & Weende          & ALTUM     & 5.1   & 4   & 12 & 16693  & train/val \\ 
                Tr14 & 2022-03-31 & Reinshof        & ALTUM     & 9     & 1   & 11 & 9882   & train/val \\ 
                Tr15 & 2022-03-31 & Reinshof        & ALTUM     & 18.5  & 1   &  8 & 6104   & train/val \\ 
                Tr16 & 2022-03-31 & Holtensen II    & ALTUM     & 3.4   & 1   & 21 & 35672  & train/val \\ 
                \midrule
                 & & & & & & total & 902121 & train/val \\
                 & & & & & & & 128550 & test \\
            \bottomrule
        \end{tabular*}   
    \end{center}
    \caption{Overview of datasets. All images are given as 5-channel multispectral data with $\SI{144}{\pixel}\times\SI{144}{\pixel}$ of size.}
    \label{tab:data_overview} 
\end{table*}

\begin{table}[t]
    \small
    \begin{center}
        \begin{tabular*}{\columnwidth}{l@{\extracolsep{\fill}}l@{\extracolsep{\fill}}l@{\extracolsep{\fill}}l@{\extracolsep{\fill}}}\toprule
            \multicolumn{2}{c}{\textbf{band name}} & \multicolumn{2}{c}{\textbf{sensor}} \\
             long & short & ALTUM & RedEdge \\\midrule
             blue & B & \SIrange{459}{491}{\nano\meter} & \SIrange{465}{485}{\nano\meter} \\
             green & G & \SIrange{548}{572}{\nano\meter} & \SIrange{550}{570}{\nano\meter} \\
             red & R & \SIrange{661}{675}{\nano\meter} & \SIrange{663}{673}{\nano\meter} \\
             red edge & REDGE & \SIrange{711}{723}{\nano\meter} & \SIrange{712}{722}{\nano\meter} \\
             near infrared & NIR & \SIrange{814}{870}{\nano\meter} & \SIrange{820}{860}{\nano\meter} \\
             thermal & TH & \SIrange{8}{14}{\micro\meter} & - \\
            \bottomrule
        \end{tabular*}   
    \end{center}
    \caption{Spectral ranges of the two camera sensor systems used in this work. The thermal band is not used in this work.}
    \label{tab:gsd_bands} 
\end{table}

All experiment fields are equipped with weather sensors, allowing for hourly temperature and humidity measurements in the fields. We can use this data to infer some more environmental quantities. We particularly focus on two of them. Firstly, a basic yet widely used quantity in phytology that connects the local weather with the development stage of the crop is the cumulative \acp{GDD}. For each day, the plant accumulates a so-called \textit{thermal sum} calculated by
\begin{align}
    \ac{GDD} = \frac{\SI{1}{\day}}{24} \sum\limits_{t=0}^{23} \left(\frac{T_\text{max}(t) + T_\text{min}(t)}{2} - T_\text{base}\right)\,.
\end{align}
For the hourly maximum and minimum temperatures $T_\text{max}(t)$ and $T_\text{min}(t)$ additionally applies an upper and lower bound
\begin{align}
    T(t) =
    \begin{cases}
        T_\text{max} & ,\, T(t) > T_\text{max}\\
        T_\text{base} & ,\, T(t) < T_\text{base}\\
        T(t) & \text{else}
    \end{cases}\,,
\end{align}
where the base and maximum temperatures $T_\text{base}$ and $T_\text{max}$ are plant-specific parameters. For sugar beet, it is empirically shown that $T_\text{base}=\SI{1.1}{\celsius}$ and $T_\text{max}=\SI{30}{\celsius}$.~\cite{gdd_formula}. The cumulative quantity of \acp{GDD} beginning at the sowing date is, after all, a proxy of the plant's development.
Secondly, we can calculate a disease-specific quantity. Simply put, the time between the infection of a plant with Cercospora and the ability of infecting other plants is called a \textit{generation} or \textit{incubation period}. The thermal sum of one incubation period for Cercospora in sugar beet is found to be \SI{4963}{\celsius\times\hour} with $T_\text{base}=\SI{6.3}{\celsius}$ and $T_\text{max}=\SI{32}{\celsius}$~\cite{bleiholder_1}. For each hourly summand, there is an additional empirical correction coefficient based on the relative humidity: the hourly summand is multiplied by $\frac{7}{8}$ if the hourly relative humidity is less than \SI{80}{\percent}. If it is at least \SI{80}{\percent}, the summand is multiplied by $\frac{9}{8}$.~\cite{npg_formula,bleiholder_2} Summation of the thermal sum and division by the incubation period yields a quotient, that describes the potential number of incubation periods a hypothetically infected plant could have undergone. We call this the \ac{NPG}.
Thus, given environmental information, we can calculate field- and recording-date-specific parameters that can serve as additional data to support the individual plant image data.

\begin{figure}[!htb]
	\centering
	\includegraphics[width=\plotwidth]{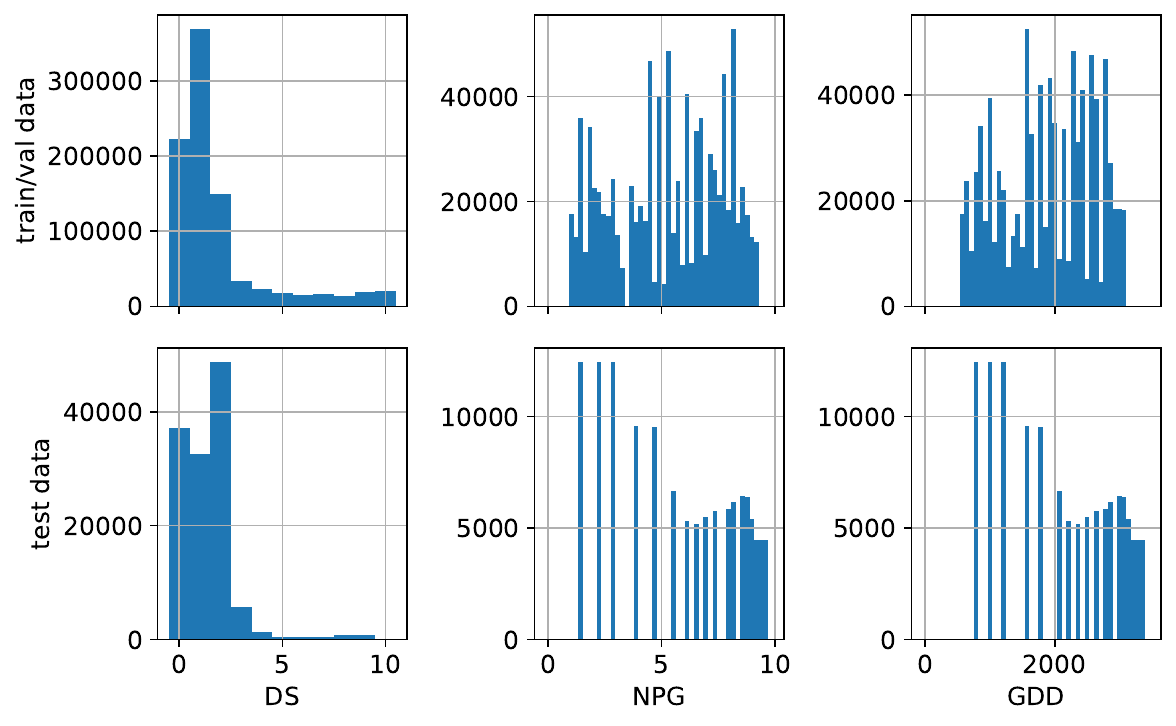}
	\caption{Histograms of available labels for \ac{DS}, \ac{NPG}, and \ac{GDD} separated by train/validation and test data.}
	\label{fig:label_abundance}
\end{figure}

\subsubsection{Image Normalization}\label{sec:normalization}

\begin{figure}[!htb]
    \centering
    \begin{subfigure}{\plotwidth}
        \includegraphics[width=\plotwidth]{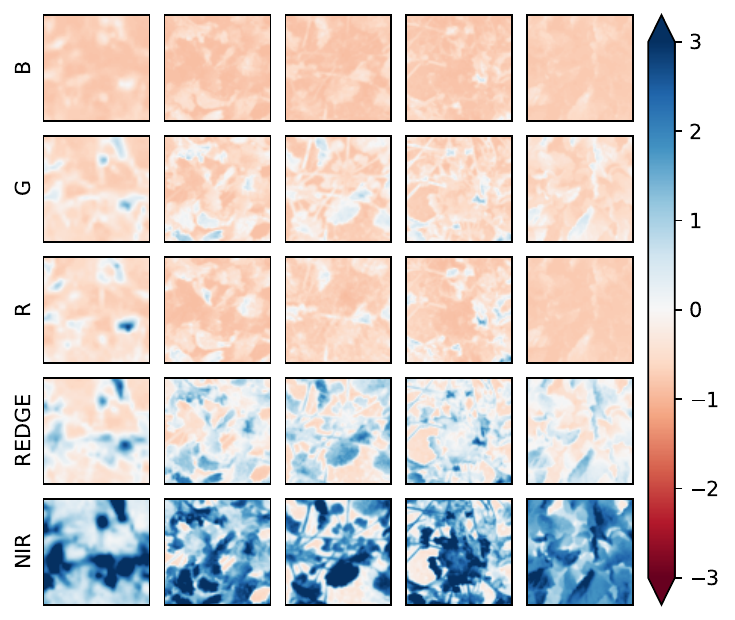}
            \subcaption{Total standardization.}
    \end{subfigure}
    \hfill
    \begin{subfigure}{\plotwidth}
        \includegraphics[width=\plotwidth]{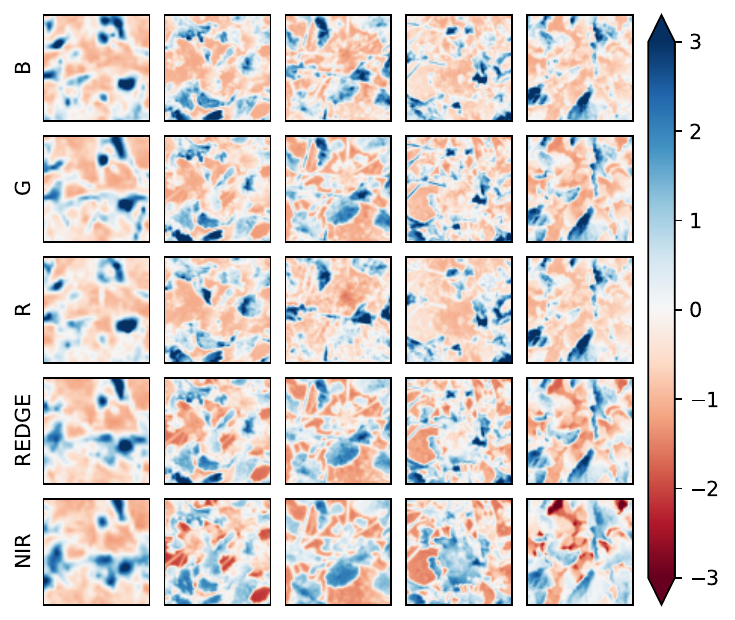}
            \subcaption{Channel-wise standardization.}
    \end{subfigure}
    \caption{Example images shown by its separate channel components and processed with total and channel-wise standardization, respectively.}
    \label{fig:standardize_tot_ch}
\end{figure}

In the vast majority of machine learning tasks dealing with image processing, images are normalized to ensure interoperability and robustness against varying image recording conditions. Additionally, numerical issues in forward and backward pass to deep learning architectures lead to the usage of data values around zero. A naive yet common approach is a simple standardization of the image data by subtracting the channel-specific mean $\mu_c$ or a global mean $\mu$ and division by the \ac{std} $\sigma_c$ or $\sigma$, respectively, for each image channel $\mathbf{C}$ in the image $\mathbf{I}$. The standardization can be done with precalculated (channel-wise) means and \acp{std} by using the information of the whole given dataset, with image specific means and \acp{std}, or even with fixed, suggested values. In this work, we will assume that our reflectance images dataset could possibly have a bias. Therefore, we standardize each image by only using its own information. Further, we differ between channel-wise and a total standardization by using means and \acp{std} for each channel separately and cross-channel calculations, respectively. Thus, we get channel-wise standardized images $\mathbf{I}_s^\text{ch}$ and total standardized images $\mathbf{I}_s^\text{tot}$ by
\begin{align}
	\mathbf{I}_s^\text{ch} &= \left\{\frac{\mathbf{C}-\mu_c}{\sigma_c}\right\}_{\mathbf{C}\in\mathbf{I}}\,,\\
        \mathbf{I}_s^\text{tot} &= \frac{\mathbf{I}-\mu}{\sigma}\,.
\end{align}
Figure~\ref{fig:standardize_tot_ch} shows some example images separated into its channel components and normalized with those two standardization methods.

A more sophisticated normalization method that, however, comes with more computational effort, is to make use of the image histogram, i.e. the abundance of data values in the images. The \ac{HE} is a contrast enhancement method that is broadly used in computer vision and image processing task for many application fields like medical imaging, as well as for signal processing like in speech recognition~\cite{hist_eq_study}. Briefly spoken, with respect to images, particularly, the idea is to normalize the elementary pixel values by their abundance. Thus, the number of pixels in each bin, or range of contrast, is equalized. As a result, each image is forced to use the full range of possible contrast.

Generally, there are two basic methods\textemdash local and global. Unlike global methods, local methods additionally use the environment of the corresponding pixel for equalization. They are usually grouped under the term \textit{Adaptive Histogram Equalization} (AHE) where a prominent method is called \textit{Contrast Limited AHE} (CLAHE)~\cite{ahe}. The adaptive methods are used for image processing tasks, where the pure contrast between neighboring objects is important, like in medical applications for tomography images~\cite{clahe_tomography}. In this work, we apply the \ac{HE} method and introduce a channel-wise and cross-channel variant analogous to the standardization. For the histogram-based methods, a lower and upper limit has to be predefined to which the values are scaled. In this work, we chose the values to be in the range $[-1,1]$.  

\subsubsection{Data Augmentation}\label{sec:data_augmentation}

A common approach to artificially increase the dataset size is data augmentation. For image data, the principle is to sample similar images around \enquote{real} data instances by various techniques like flip, rotation, color and brightness jitter, etc. Obviously, different use cases allow for different augmentation methods. For instance, in case of medical imaging task, one is mostly bound to the image orientation. In case of street scene images for autonomous driving applications, a vertical flip, i.e. put the image upside down, does not make any sense. However, both cases eventually allow for changes in brightness and/or contrast. In our case of plant images, we fortunately have all degrees of freedom regarding flips and rotations. Thus, we flip each image randomly with a probability of \SI{25}{\percent} and rotate each image by a random angle. Brightness and contrast jitters are not necessary, since our normalization methods neutralize them. Additionally, we can exploit this principle in model inference mode by evaluation the images in different rotations and average the predictions. In order to be robust against different ground sampling distances and accompanied resolution changes, we introduce a Gaussian blur augmentation. With a probability of \SI{10}{\percent}, an image is blurred at a strength of \SIrange{3}{8}{\pixel}. Another optional augmentation, is a random channel dropout. With a probability of \SI{25}{\percent}, we drop the information of up to \num{3} channels. Although it is quite unlikely, that in the application, single channels will be dropped, it is interesting to train models being robust against missing information in order to see, how important each image channel is for the prediction of our target quantity. With the channel dropout, we lower the ability of the model to focus on single channels and rather connect information among all channels.

Next, we shed light on one purpose the data has been recorded for and how we will use it for the use case described in this work. 

\subsubsection{Use Case: Disease Severity Estimation}\label{sec:use_case_description}

The goal behind our use case in this work is about determining the \ac{DS} in \ac{CLS}-infected sugar beet fields, one of the most damaging foliar diseases in sugar beet cultivation. \ac{CLS} is caused by the fungus \textit{Cercospora beticola}. Symptoms appear as numerous small, round, gray spots with a red or brown border on leaves. As the infestation increases, the leaves become necrotic and dry up. When a large part of the leaf area is lost, the plant often tries to recover by generating a new birth of leaves at the cost of its stored sugar. However, if the conditions are favorable for the fungus and the attack is very severe, the plants die.~\cite{cls_1,cls_2} The observation of plant diseases is usually done by visual scoring. Being the visual field scoring of \ac{DS} an activity that requires a lot of time and well-trained personnel~\cite{disease_assessment} is the main bottleneck in the control of \ac{CLS}. Therefore, it is desirable to have an automatic \ac{DS} estimation model.

Considering heterogeneous disease distribution within sugar beet fields, a detailed and geo-referenced assessment of \ac{DS} might lead to precise protection measures within the canopy. Geo-referenced and plant-based determination of \ac{DS} is therefore essential. A naive approach for the prediction of \ac{DS} with single plant images would be to model a classification problem like in~\cite{ds_classification}. Despite being a valid approach at first sight, certain phenotypical knowledge enables to model this problem more intelligently. In the following sections, we will explain our basic paradigm to solve the \ac{DS} estimation problem and our proposed deep learning model based on it.

\begin{figure*}[!htb]
	\centering
	\includegraphics[width=\textwidth]{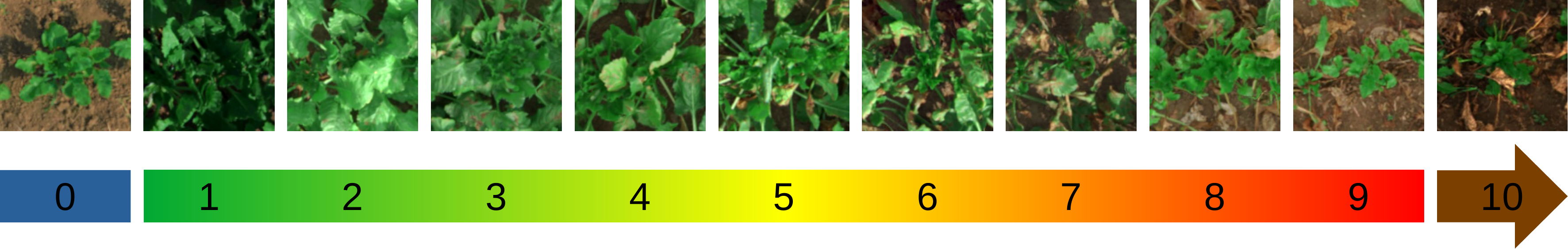}
	\caption{Used disease severity scale for our prediction model with example images. The scale is based on the usual \ac{CLS} rating scale. We added the 0 for non-infested sugar beets before canopy closure, and the 10 for newly sprouted plants as in~\cite{facu_sugarindustry}.}
	\label{fig:ds_scale}
\end{figure*}

First, we have to define a \ac{DS} annotation scale that serves as a guideline for all human expert annotations and finally as \enquote{unit} of the model input. In this work, the rating scale developed in~\cite{facu_sugarindustry} will be used with an extension for non-infested and newly sprouted plants. Figure~\ref{fig:ds_scale} shows the numerical scale with exemplary plant images. The scale from 1 to 9 belongs to definitions of the KWS scale. The KWS scale is a severity diagram that ranges from 1 to 9. A rating of 1 indicates the complete absence of symptoms, while a rating of 3 indicates the presence of leaf spots on older leaves. A rating of 5 signifies the merging of leaf spots, resulting in the formation of necrotic areas. A rating of 7 is assigned when the disease advances from the oldest leaves to the inner leaves, leading to their death. Finally, a rating of 9 is given when the foliage experiences complete death~\cite{cercospora_kws}.

In field experiments, we often face data or annotations that need a high effort to acquire. Nevertheless, several data can be acquired rather automatically or with low human effort. In this work, we call this \enquote{cheap} and \enquote{expensive} labels. The \ac{DS} acquisition is rather expensive, while, for instance, weather data acquired with automatic sensors or public weather stations is, typically, relatively cheap. Additionally, the development and epidemiology of the pathogen and disease \ac{CLS} is highly influenced by specific environmental conditions.~\cite{fungal_leaf_disease_management} In this work, we will make use of the cheap data in order to increase efficiency on expensive data. As shown above in Section~\ref{sec:field_data}, we have the weather-based parameters \ac{GDD} and \ac{NPG}. They are not plant specific but, at least, specific for the recording date. Thus, we can annotate many plants with those labels at one stroke. Those labels are, surely, not as meaningful as manually annotated labels, but they can serve for pre-training models. This is particularly interesting for our application of transformers, since they usually need lots of data: we can pretrain the model with the cheap labels and finetune on the expensive labels. Thus, a possible lower availability of the expensive labels could be compensated and training speed is enhanced if the model backbone at the start of fine-tuning stage already \enquote{knows} low-level filters and the basic concept of our input data. The two different stages of pretraining and finetuning are represented as different learning paths in our model sketch in Figure~\ref{fig:sugarvit_sketch}. Additional details of the model are discussed in the later sections of this work. First, we want to introduce in the concept behind our model architecture and, secondly, we shed light on the different model parts in detail.

\subsection{Deep Label Distribution Learning}

If classification problems can be formulated within an ordinal scale, the transfer into a regression task might be a good choice. However, if the classification is very granular, the collection of data with precise labels can be challenging. Rather than learning distinct, unique labels, the paradigm of \ac{LDL}~\cite{ldl} was proposed. It stabilizes the model training of labels by modeling their ambiguity. It is used for tasks like facial age estimation~\cite{ldl_age} or head pose estimation~\cite{ldl_pose}. In combination with deep neural networks, the paradigm is referred to as \ac{DLDL}~\cite{dldl}. In \ac{DLDL}, the output of a deep neural network mimics the label distribution by a series of neurons that learn a discrete representation of the probability density function. This representation is commonly known as the \ac{pmf}. Thus, the labels have the form of a probability distribution and the obvious difference in contrast to a pure regression is that the network output is not only based on a single neuron, whereas the difference to a pure classification is that, in contrast to one- or multi-hot-labels, also neighboring neurons are triggered which stabilizes regions, where fewer data is available.
Two additional advantages, especially for the use case in this work, are, firstly, that we can easily model uncertainty of labels. The \ac{DS} annotation is based on individual human experts' judgement. Often, different plants are annotated by different experts, which causes uncertain classifications. Secondly, the model output becomes more transparent, since one can observe how confident the model is in its prediction by comparing the shapes of true and predicted label distribution. Thus, \ac{DLDL}, once more, is an ideal way to model these annotations.

\subsubsection{Full Kullback-Leibler Divergence Loss}

The \ac{DLDL} approach proposed in~\cite{dldl} utilizes a L1 loss for the expectation value and a \ac{KLD}~\cite{kl_div} loss for the label distribution. However, L1 and \ac{KLD} loss originate from different statistical concepts and, therefore, have scales that are, per se, not comparable. In most cases, a weighting parameter has to be introduced, resulting in an artificial hyperparameter of the model. Our approach circumvents the problem by reformulating the L1 loss as a \ac{KLD} loss. Additionally, we further accelerate the training by introducing a \enquote{smoothness} regularization to the label distribution. The regularization is also formulated as a \ac{KLD} loss, not needing any hyperparameter, either. Furthermore, the gained scale invariance not only makes the components comparable, but also enables the cross-comparability between different labels. This is especially interesting in the use case of this work, since we aim a joint regression of diverse phenological parameters, probably having different domains. This novel loss function is already introduced in~\cite{full_kl_loss}. However, since the approach is very well suited for the use case in this work, we will introduce the 3 loss components again in the following.

\subsubsubsection{Label Distribution Loss}\label{sec:ldl_loss}

Let $\mP(y\mid\bx)$ be the true label distribution for a given data point, i.e. an image, $\bx$. Then, the label distribution loss $L_{ld}$ is the discrete Kullback-Leibler divergence between the true and predicted label distribution,
\begin{align}
	L_{ld} &= \KLD(\mP||\hat{\mP}) = \sum_{y} \mP(y\mid\bx)\log\frac{\mP(y\mid\bx)}{\hat{\mP}(y\mid\bx)}\,,\label{ldl_loss}
\end{align}
where the hat denotes the prediction. This definition follows the label distribution loss given in~\cite{dldl}.

\subsubsubsection{Expectation Value Loss}

Unlike in~\cite{dldl}, we formulate the expectation value loss as a \ac{KLD} of truth and prediction as if both of them were normal distributions $\mathcal{N}(\cdot\mid\mu,\sigma^2)$ with expectation value $\mu$ and variance $\sigma^2$. For the model predictions, $\hat{\mu}$ and $\hat{\sigma}^2$ can be calculated via
\begin{align}
	\hat{\mu} = \sum_y y \hat{\mP}(y\mid\bx)\,,\quad\hat{\sigma}^2 = \sum_y (y-\hat{\mu})^2 \hat{\mP}(y\mid\bx)\,.
\end{align}
Thus, our expectation value loss is
\begin{align}
	L_{exp} &= \KLD(\mathcal{N}(\cdot\mid\mu,\sigma^2)||\mathcal{N}(\cdot\mid\hat{\mu},\hat{\sigma}^2))\nonumber\\
	&= \log\frac{\hat{\sigma}}{\sigma} - \frac{1}{2} + \frac{\sigma^2 + (\hat{\mu} - \mu)^2}{2\hat{\sigma}^2}\,.
\end{align}
Detailed calculation steps can be found in the Appendix of~\cite{full_kl_loss}.

\subsubsubsection{Smoothness Regularization Loss}
In order to accelerate the training process, especially in early stages, we force the predicted label distribution to be smooth by a \ac{KLD} regularization term. The idea is to shift the predicted distribution by one position which we call $\hat{\mP}^s$ and calculate a symmetric discrete \ac{KLD}, i.e. we average both shift directions. Thus,
\begin{align}
	L_{smooth} &= \frac{1}{2} \left[ \KLD(\hat{\mP}||\hat{\mP}^s) + \KLD(\hat{\mP}^s||\hat{\mP})\right]\nonumber\\
	&= \frac{1}{2} \sum_y (\hat{\mP}(y\mid\bx) - \hat{\mP}^s(y\mid\bx)) \log\frac{\hat{\mP}(y\mid\bx)}{\hat{\mP}^s(y\mid\bx)}\,.
\end{align}

Finally, a sum combines the loss components. Thus, our final loss is
\begin{align}\label{eq:loss_comp}
	L = L_{ld} + L_{exp} + L_{smooth}\,.
\end{align}

\subsubsection{Multi-Head Regression}\label{sec:multihead_reg}

If multiple sources of labels are available, it may be considerable to perform the regression with multiple labels jointly. Each regression problem is then realized by an own so called \enquote{regression head}, i.e., a sub-model, that is trained to transform the feature representation from the backbone into the respective label space of interest. Especially for large backbone models, this has the advantage that only one backbone is needed for multiple purposes, which reduces the total model size. We further call this concept \enquote{Multi-Head Regression}.

\subsection{Model Architecture}

\begin{figure*}[!htb]
	\centering
	\includegraphics[width=\textwidth]{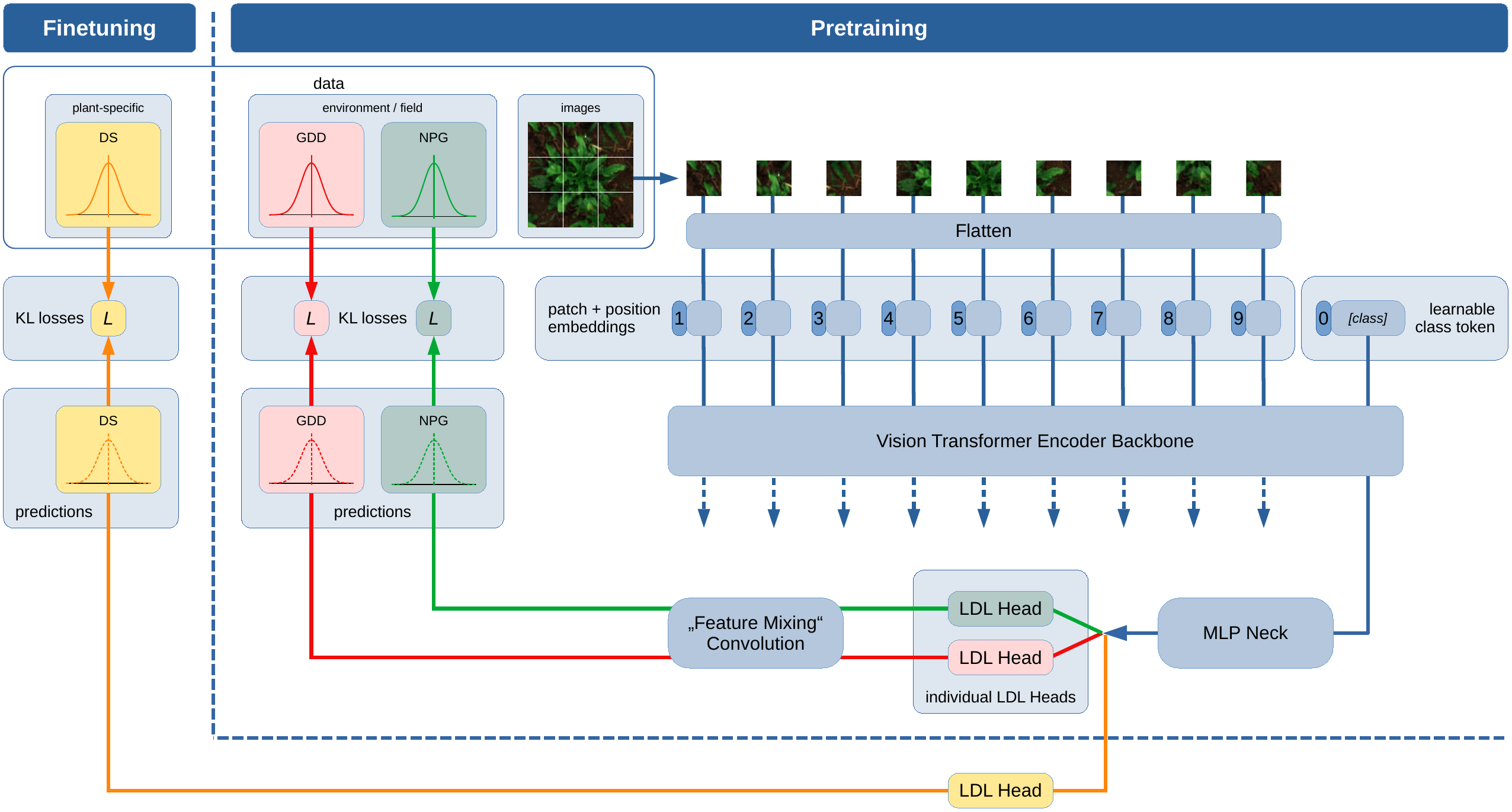}
	\caption{Sketch of our proposed Multi Deep Label Distribution Learning (Multi-\ac{DLDL}) network with a \acf{ViT} backbone. The \acf{LDL} heads are trained with separate optimizers and loss functions. The \ac{ViT} and \ac{MLP} part are the joint basis and are trained in each backward pass of the \ac{LDL} heads. As output of the \ac{ViT}, the last hidden state of the learnable class token is used. Furthermore, our use case is shown by having multispectral plant image data and two training stages. The pretraining is done on the environmental, field-related quantities \acf{GDD} and \acf{NPG}. The target label \acf{DS} is trained in the subsequent finetuning stage. In principle, the model can be generalized to more labels in each training stage by adding more \ac{LDL} heads.}
	\label{fig:sugarvit_sketch}
\end{figure*}

In this chapter, we shed light on the architecture of our proposed model. Figure~\ref{fig:sugarvit_sketch} shows all the building blocks of our proposed model, further called \enquote{\sugarvit{}}. We now describe the 3 main building blocks of \sugarvit{} and its design motivations.

\subsubsection{Vision Transformer Backbone}

In recent years, transformer architectures are successfully utilized for diverse deep learning tasks. Especially in the field of \ac{NLP}, transformer-based models show great success.~\cite{transformers_in_nlp}. In \ac{NLP}, transformers learn structures in sequential data like text or sentences by processing its basic building blocks, commonly knows as \enquote{tokens}. To use this principle also for classification tasks on image data, the Vision Transformer (\ac{ViT}) model has been proposed.~\cite{vit} The main principle is to divide an input image into flattened tiles that are processed by several multi-head attention layers~\cite{attention}. An additional learnable \enquote{class token} is added, which processed output is passed through a classification head. Figure~\ref{fig:sugarvit_sketch} also shows the mentioned building blocks. By comprising many attention blocks and hidden layers, (vision) transformer architectures are complex and need large amounts of data. Thus, they are usually pretrained on large-scale datasets like ImageNet~\cite{imagenet} for most of the image processing tasks. For the use case this work is about, we process multispectral 5-channel images rather than \ac{RGB} images. Therefore, we cannot use ImageNet-pretrained architectures per se. However, the plant image dataset used in this work is large enough to train an architecture with a vision transformer backbone from scratch, as we will see in further sections. The goal of the learning process is that the \ac{ViT} backbone is trained to be an expert in understanding the image as a whole and extract remarkable traits to \enquote{encode} the image information into a rich feature space.

\subsubsection{MLP Neck}\label{sec:mlp_neck}

In the original \ac{ViT} model, an \ac{MLP} is used as a classification head. Since we want to build a multi-head regression model (cf. Section~\ref{sec:multihead_reg}), we use an \ac{MLP} as an intermediate layer between the \ac{ViT} backbone and the regression heads. If certain labels have something in common or share some information, i.e., latent label correlations, this \enquote{neck} sub-model between backbone and heads is trained to learn those latent correlations. We could exploit this principle in our use case by introducing a simple \enquote{cheap} feature, i.e., that is easy to measure and has a more or less obvious correlation to the \ac{DS}. For instance, we could choose the interval between the image recording date and the date where canopy closure can be observed in the corresponding field experiment. Canopy closure means that neighboring plants touch each other, resulting in a closed field vegetation. In the following, we call this feature \ac{DAC}. Obviously, this feature has negative values before canopy closure is reached. Alternatively, one could also think about including the days after sowing. The \ac{DAC} are expected to guide the model to the correct \ac{DS} by having some correlation with it, e.g., young plants (low \ac{DAC}) are probably less severely infected, whereas older plants (high \ac{DAC}) are probably rather severely infected. In addition, the infection probability raises when the plats are in contact. All in all, the \ac{MLP} neck part is a part of the model, where expert knowledge and known correlations can be integrated. Please note here that in the experiments, we follow another approach to integrate associated knowledge in the model to predict the \ac{DS}. Nevertheless, the above approach can be a valid, as well.

\subsubsection{LDL Heads}

For each label, the output of the \ac{MLP} neck is processed by a separate, so called, \ac{LDL} head. It consists of individual \acp{FCL} after the \ac{MLP} neck for each label. The idea behind the individual networks is to enable the model to learn label-individual transformations from the cross-label output of the \ac{MLP} neck. Thus, the \ac{LDL} heads are trained to be experts in their label domain and be able to transform the feature space to their regarding label space. When passed through these layers, the features are mixed with a component we call \textit{Feature Mixing}.

\subsubsection{Feature Mixing}

In the Feature Mixing component, the output layers of all individual \ac{LDL} Heads are combined linearly. This enables the model to scale and mix information of other labels into specific labels. The mixing coefficients can be learned during training and are initialized to the unit matrix, i.e., at training start, only the respective label is used. A final \ac{FCL} for each label maps the mixed features into the corresponding label distribution space. The size of the \ac{FCL} is determined by the number of discretization or quantization steps that can be different for each label. A softmax activation ensures the outputs of the \acp{FCL} sum to 1 each. Thus, the \ac{FCL} approximates the label distribution in the form of a \ac{pmf}. On this \ac{pmf}, we can then evaluate our \ac{KLD} loss functions and, finally, train with the ground truth label distributions.

\section{Results}\label{sec:results}

In this section, we will introduce our performed experiments. At first, we test if the histogram equalization preprocessing step for the images is really beneficial for the model performance. 

\begin{figure}[!htb]
	\centering
	\includegraphics[width=\plotwidth]{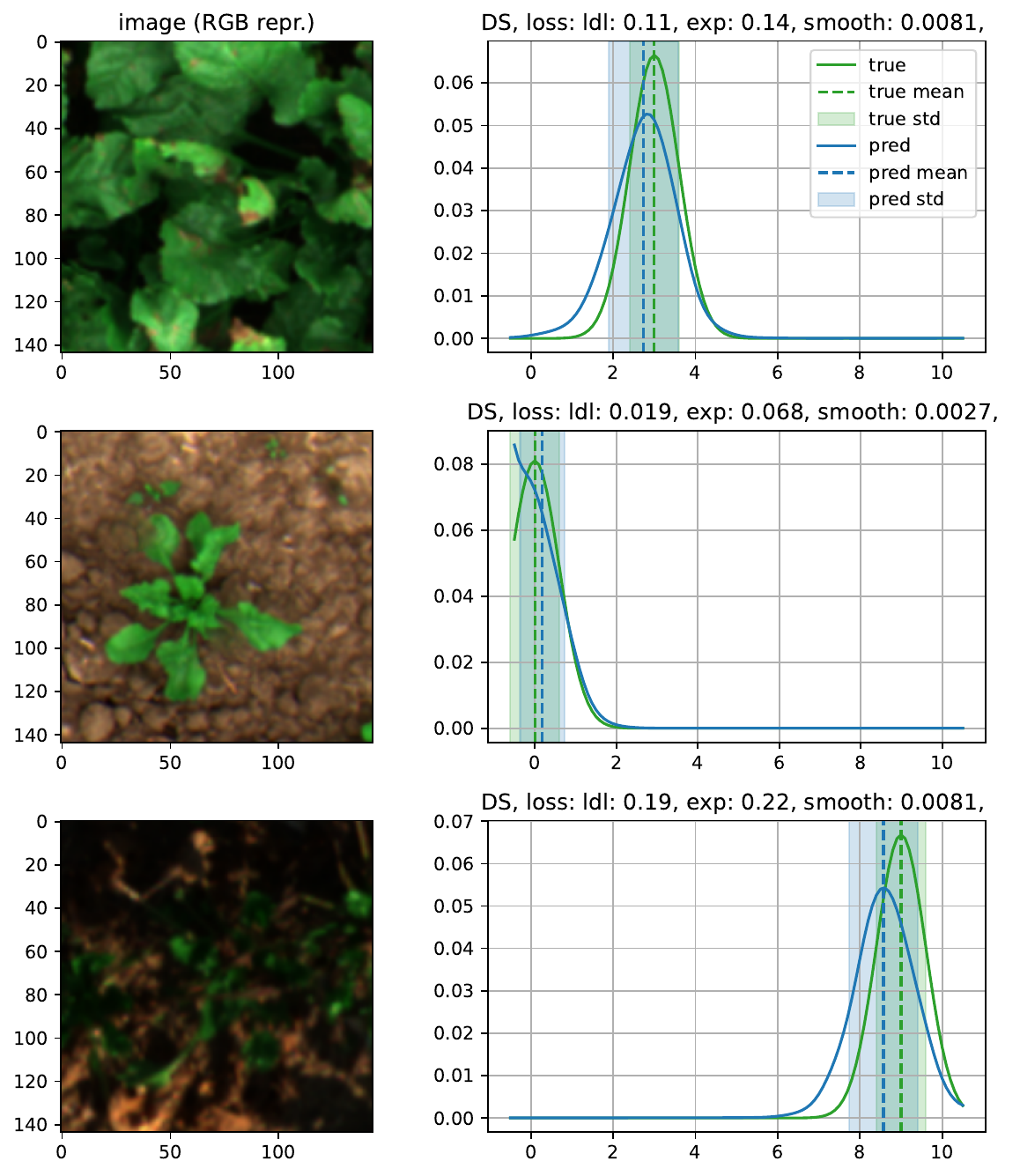}
	\caption{Output of \sugarvit{}. The \acf{DS} labels are learned as label distributions (green curves). \sugarvit{} outputs again probability distributions (blue curve). The prediction in the end is the expectation value of the output distributions (dashed lines).}
	\label{fig:example_prediction}
\end{figure}

In the next experiment, we make use of \enquote{cheap} data, i.e., weather data as mentioned in Section~\ref{sec:use_case_description}. In our case, we have weather stations in the field measuring basic weather parameters. This data is available for a whole field, thus, many single plant images. With the single images and those cheap labels, we can perform a pretraining of the backbone. However, this is another approach to combine cheap and expensive data than the one mentioned in section~\ref{sec:mlp_neck}, the major advantage is that in the final model, only the label of interest is used, which results in a slightly lighter model and decreases inference time. After pretraining, we perform a model training with \ac{DS} labels, resulting in our \sugarvit{} model. Example predictions of a trained \sugarvit{} model are shown in Figure~\ref{fig:example_prediction}.

In a last experiment, we finally investigate if the pretraining also improves the performance of \sugarvit{} by comparing the finetuned \sugarvit{} with a one only trained by the \ac{DS} labels. We further compare a non-pretrained \sugarvit{} model to a one that is only trained on with \ac{RGB} bands information in order to see, whether the beyond-optical spectral information is important. 

Before performing the actual experiments, the variances for the \ac{DS} label distributions must be set, since there is no individual information for each data point, or image. In this work, we model the \ac{DS} label distributions by normal distributions $\mathcal{N}(\cdot\mid\mu,\sigma^2)$ with the experts' labels as expectation values $\mu$ and a variance $\sigma^2$ that is based on an assumed standard error. We set $\sigma_{DS} = \num{0.6}$ as a \enquote{human estimation} error. Please note, that this is no empirically found error but rather an educated guess.

In all experiments, we randomly split the training and validation data (cf. Table~\ref{tab:data_overview}) with initialization seeds assuring reproducibility. For our dataset, the plants have mostly low \ac{DS} scores in images of the early plant growth, leading to label imbalance, as seen in the histograms in Figure~\ref{fig:label_abundance}. Looking at Table~\ref{tab:data_overview}, the sizes of the datasets are quite different. In order to minimize the bias and to prevent the model from focusing on few labels and datasets, we use a weighted sampling of the data. The weight of each image is the inverse of the total abundance of its \ac{DS} label times the size of the respective dataset. Thus, in each training batch, the distribution of datasets and labels is uniformly distributed in average.

Finally, we define a validation metric. Since our model outputs distributions, metrics like root mean squared error or \ac{MAE} are not appropriate since they do not give information about the overall distribution. Alternatively, we use the mean overlap between the predicted and true label distribution. Since the distributions are \acp{pmf}, the calculation of the \ac{MDO} for a batch of $N$ instances is
\begin{align}
    \ac{MDO} = \frac{1}{N}\sum_{i=1}^N \sum_y \left\{\min\left(\mP(y\mid\bx), \hat{\mP}(y\mid\bx)\right)\right\}_y\,.
\end{align}
The \ac{MDO} takes values between $0$ and $1$ where $1$ indicates perfect overlap. For the validation, we use the same weighted sampling as in the training stage, to validate on pseudo-uniform distributed labels. Thus, we respect the prediction quality for each label equally and independent of the total label abundance in the dataset.

\subsection{Standardization vs. Histogram Equalization}\label{sec:norm_exp}

Before we perform the training of our \sugarvit{} model, we evaluate how the histogram equalization improves the model performance in favor of a \enquote{simpler} standardization preprocessing method. To have a potentially more universal model in the event, we do not assume that we know the dataset as a whole. Thus, we use normalization only based on the information of a single image, as already mentioned in Section~\ref{sec:normalization}. In the following experiment, we compare the standardization and the histogram equalization method, respectively, with channel-wise and cross-channel calculation. For each method, we perform \num{10} runs with different initialization seeds with the model configuration given in Table~\ref{tab:normtest_modeldetails} and trained for \num{80} epochs. To speed up the training, we only use the datasets Tr01, Tr02, and Tr03 (cf. Table~\ref{tab:data_overview}). Those are randomly split into a training and validation subset by ratio \SI{80}{\percent}:\SI{20}{\percent}.

\begin{table}[t]
	\footnotesize
	\begin{center}
		\begin{tabular*}{\columnwidth}{l@{\extracolsep{\fill}}r}\toprule
			\textbf{ViT backbone} & \\\midrule
                input size & $\num{5}\times\SI{144}{\pixel}\times\SI{144}{\pixel}$ \\
                patch size & $\SI{12}{\pixel}$ \\
                hidden size & $\num{512}$ \\
                \# hidden layers & $\num{4}$ \\
                \# attention heads & $\num{4}$ \\
                intermediate size & $\num{512}$ \\
                activation hidden layers & GELU \\
                dropout hidden layers & \num{0.02} \\
                dropout attention & \num{0.02} \\
			\midrule
			\textbf{MLP neck} & \\\midrule
			layer size & \num{512} \\
			layers & \num{3} \\
			activation & ReLU \\
			dropout & \num{0.2} \\
			\midrule
			\textbf{LDL heads} & \\\midrule
                individual MLP layers & \num{2} \\
                individual MLP layer size & \num{256} \\
                activation & ReLU \\
                dropout & \num{0.8} \\
			DS quantization steps & \num{111} \\
			DS regression limits & [\num{-0.5}, \num{10.5}] \\
			DS label distribution std & \num{0.6} \\
			\midrule
			\textbf{Optimizer} & \\\midrule
			algorithm & AdamW \\
			initial learning rate & \num{e-3} \\
			weight decay & \num{0.1} \\
			\bottomrule
		\end{tabular*}   
	\end{center}
	\caption{Model configuration for comparison between standardization and histogram equalization.}
	\label{tab:normtest_modeldetails}  
\end{table}

\begin{figure}[!htb]
    \centering
    \begin{subfigure}{\plotwidth}
        \includegraphics[width=\plotwidth]{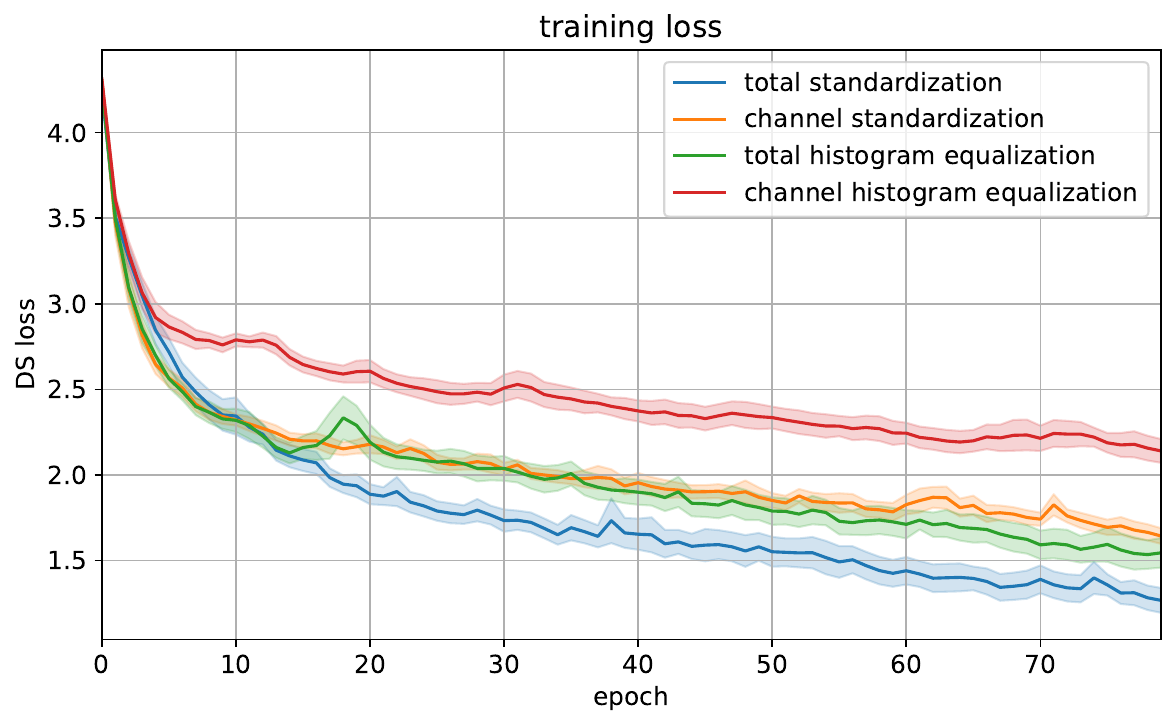}
    \end{subfigure}
    \hfill
    \begin{subfigure}{\plotwidth}
        \includegraphics[width=\plotwidth]{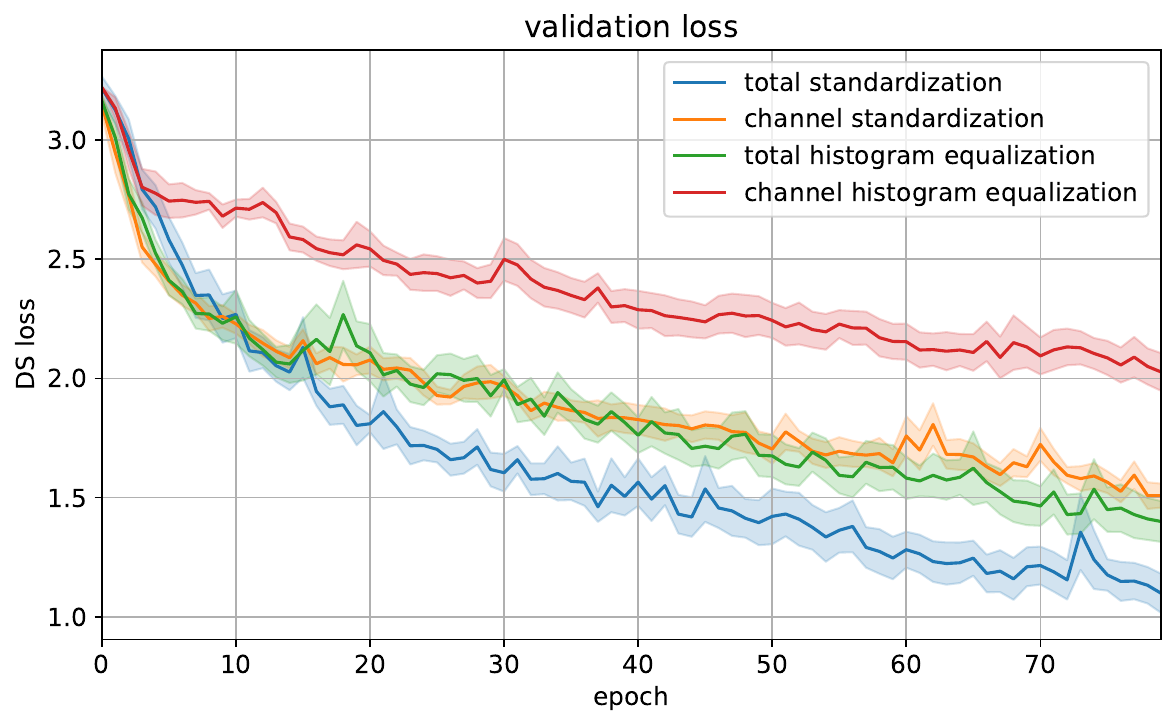}
    \end{subfigure}
    \hfill
    \begin{subfigure}{\plotwidth}
        \includegraphics[width=\plotwidth]{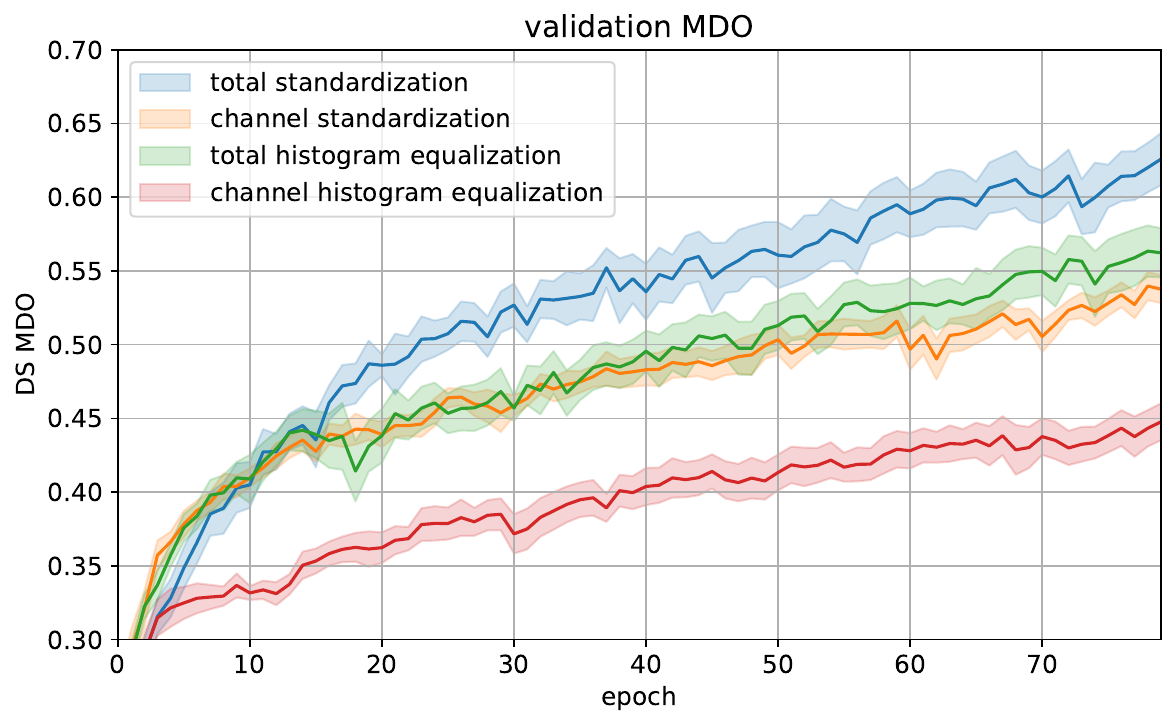}
    \end{subfigure}
    \caption{Results of the standardization vs. histogram equalization experiment. For both methods, total and channel-wise variants are shown. For each experiment, \num{10} runs with different seeds are performed. The solid lines describe the means, whereas the bands are the standard errors of the mean. The three plots show training loss, validation loss, and validation \acp{MDO}.}
    \label{fig:normtest_results}
\end{figure}

The results are presented in Figure~\ref{fig:normtest_results}. As expected, the total normalization methods perform better than channel-wise normalization. This makes sense, because for \ac{DS} prediction, an important feature is the difference in radiance of spectral bands, thus, the difference in values across channels. When normalizing the image totally by calculation of cross-channel histograms or mean and \ac{std}, respectively, this information is preserved, while in the channel-wise normalization it is lost. Nevertheless, channel-wise normalization is more robust against calibration errors of the sensor. Another remarkable observation is, that the standardization method is not only computationally more efficient than histogram equalization, but also performs better. Thus, we find the total (cross-channel) standardization method to be the best performing normalization method, and we will use this method for our \sugarvit{} model.

\subsection{\sugarvit{} Pretraining}\label{sec:sugarvit_exp}

We perform a pretraining of the \sugarvit{} model on the environmental data labels. This should prepare the model for the plant images by learning low-level features of the plant images. The configuration of our \sugarvit{} model for both pretraining and finetuning stage is listed in Table~\ref{tab:sugarvit_modeldetails}.

\begin{table}[t]
	\footnotesize
	\begin{center}
		\begin{tabular*}{\columnwidth}{l@{\extracolsep{\fill}}r}\toprule
			\textbf{ViT backbone} & \\\midrule
                input size & $\num{5}\times\SI{144}{\pixel}\times\SI{144}{\pixel}$ \\
                patch size & $\SI{12}{\pixel}$ \\
                hidden size & $\num{1024}$ \\
                \# hidden layers & $\num{8}$ \\
                \# attention heads & $\num{4}$ \\
                intermediate size & $\num{1024}$ \\
                activation hidden layers & GELU \\
                dropout hidden layers & \num{0.02} \\
                dropout attention & \num{0.02} \\
			\textbf{MLP neck} & \\\midrule
			layer size & \num{512} \\
			layers & \num{3} \\
			activation & ReLU \\
			dropout & \num{0.2} \\
			\midrule
			\textbf{LDL heads} & \\\midrule
                individual MLP layers & \num{2} \\
                individual MLP layer size & \num{256} \\
                activation & ReLU \\
                dropout & \num{0.8} \\
    		NPG quantization steps & \num{100} \\
			NPG regression limits & [\num{-0.5}, \num{11.5}] \\
			NPG label distribution std & \num{0.3} \\
    		GDD quantization steps & \num{100} \\
			GDD regression limits & [\SI{-5}{\celsius\times\day}, \SI{3500}{\celsius\times\day}] \\
			GDD label distribution std & \SI{100}{\celsius\times\day} \\
			DS quantization steps & \num{89} \\
			DS regression limits & [\num{-0.5}, \num{10.5}] \\
			DS label distribution std & \num{0.6} \\
			\midrule
			\textbf{Optimizer} & \\\midrule
			algorithm & AdamW \\
			initial learning rate & \num{5e-4} \\
			weight decay & \num{0.1} \\
			\midrule
			\textbf{Learning rate scheduler} & \\\midrule
			strategy & cyclic learning rate (linear) \\
			interval & step \\
			maximum lr & \num{1e-3} \\
			step size (up and down) & \num{500} \\
			  mode & exponential range \\
                $\gamma$ & \num{0.9999} \\			
			\bottomrule
		\end{tabular*}   
	\end{center}
	\caption{\sugarvit{} model configuration for pretraining and finetuning. For pretraining, the labels \acf{NPG} and \acf{GDD} are used. In finetuning stage, the final label of interest, \acf{DS}, is trained.}
	\label{tab:sugarvit_modeldetails}
\end{table}

\begin{figure}[!htb]
	\centering
	\begin{subfigure}{\plotwidth}
		\includegraphics[width=\plotwidth]{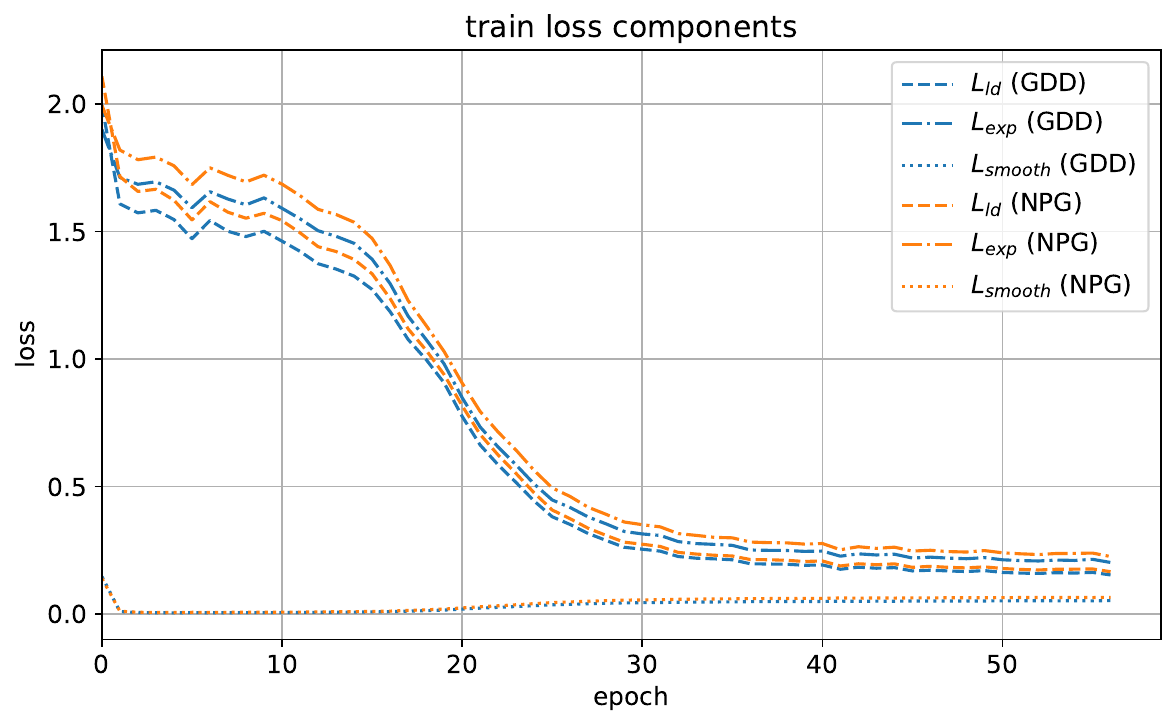}
            \subcaption{Training loss by epoch seperated into the \num{3} loss terms (cf. Equation~\eqref{eq:loss_comp}) for \ac{GDD} and \ac{NPG} labels.}
	\end{subfigure}
        \hfill
	\begin{subfigure}{\plotwidth}
		\includegraphics[width=\plotwidth]{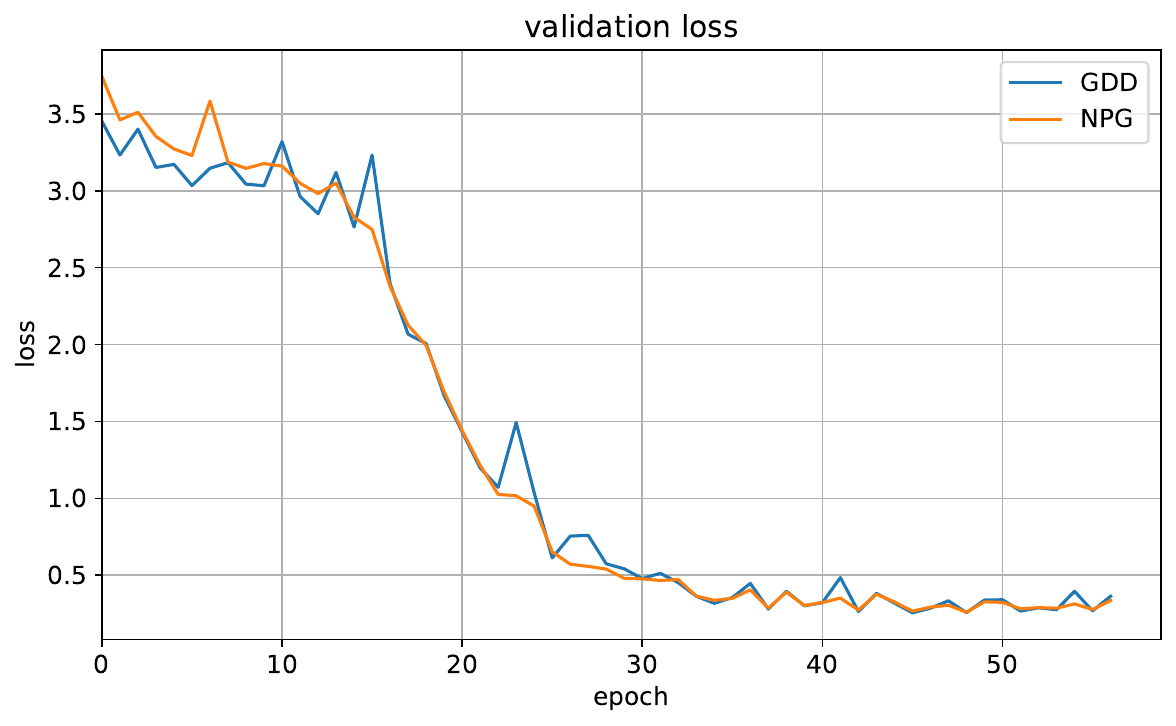}
            \subcaption{Validation loss by epoch for both \ac{GDD} and \ac{NPG} labels.}
	\end{subfigure}
        \hfill
	\begin{subfigure}{\plotwidth}
		\includegraphics[width=\plotwidth]{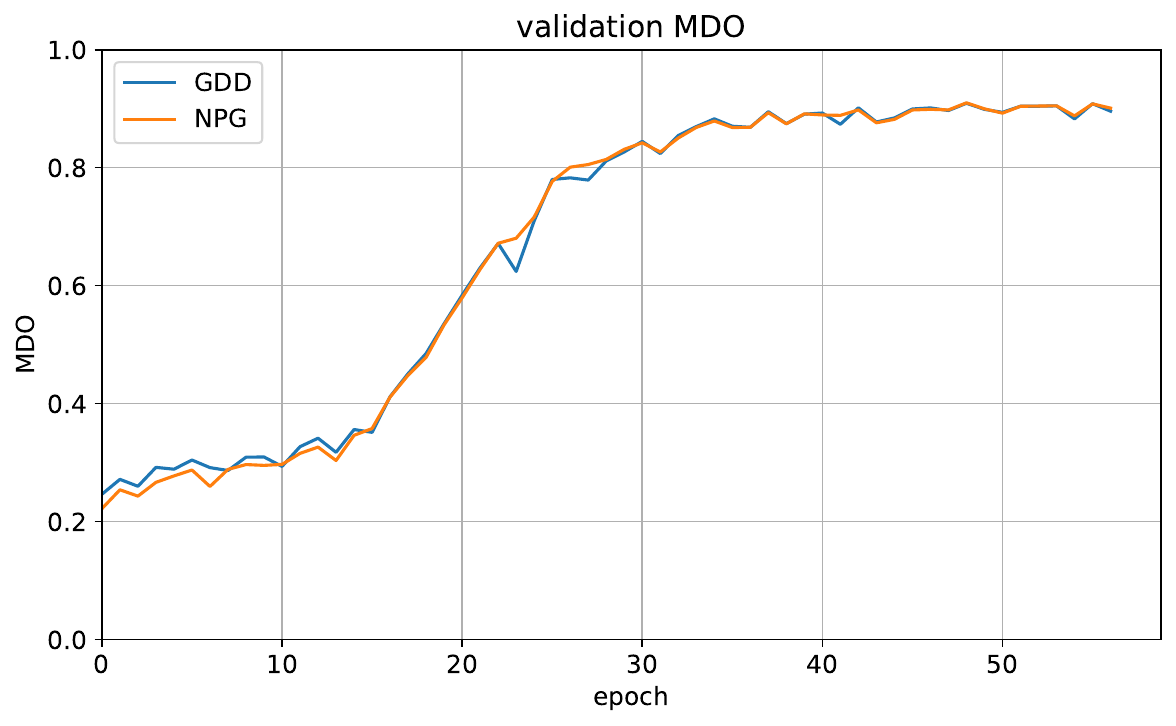}
            \subcaption{Validation \acf{MDO} by epoch for both \ac{GDD} and \ac{NPG} labels.}
	\end{subfigure}
	\caption{Results of the \sugarvit{} pretraining.}
	\label{fig:sugarvit_pretraining}
\end{figure}

The results for training and validation loss, as well as the validation \ac{MDO} are shown in Figure~\ref{fig:sugarvit_pretraining}. As seen in the training loss component plot, our loss function is indeed invariant under the label scale, as described in Section~\ref{sec:ldl_loss}. Without any scaling parameter, both \ac{GDD} and \ac{NPG} labels have a comparable loss, although the scales are very different. We see a convergence of the validation \ac{MDO} at ca. \SI{90}{\percent} after roughly \num{40} training epochs for both \ac{GDD} and \ac{NPG} labels. As the best model, we take the one with the best validation \ac{MDO} and use it for the further steps. The results for the training variants with channel dropout and \ac{RGB}-only information are very similar. Plots can be found in the supplementary material~\ref{sec:appendix}.
If the pretraining was beneficial for the subsequent finetuning regarding convergence speed and prediction quality, is shown in the following section. 

\subsection{Comparison Experiments}\label{sec:comparison_exp}

\begin{figure}[!htb]
	\centering
	\begin{subfigure}{\plotwidth}
		\includegraphics[width=\plotwidth]{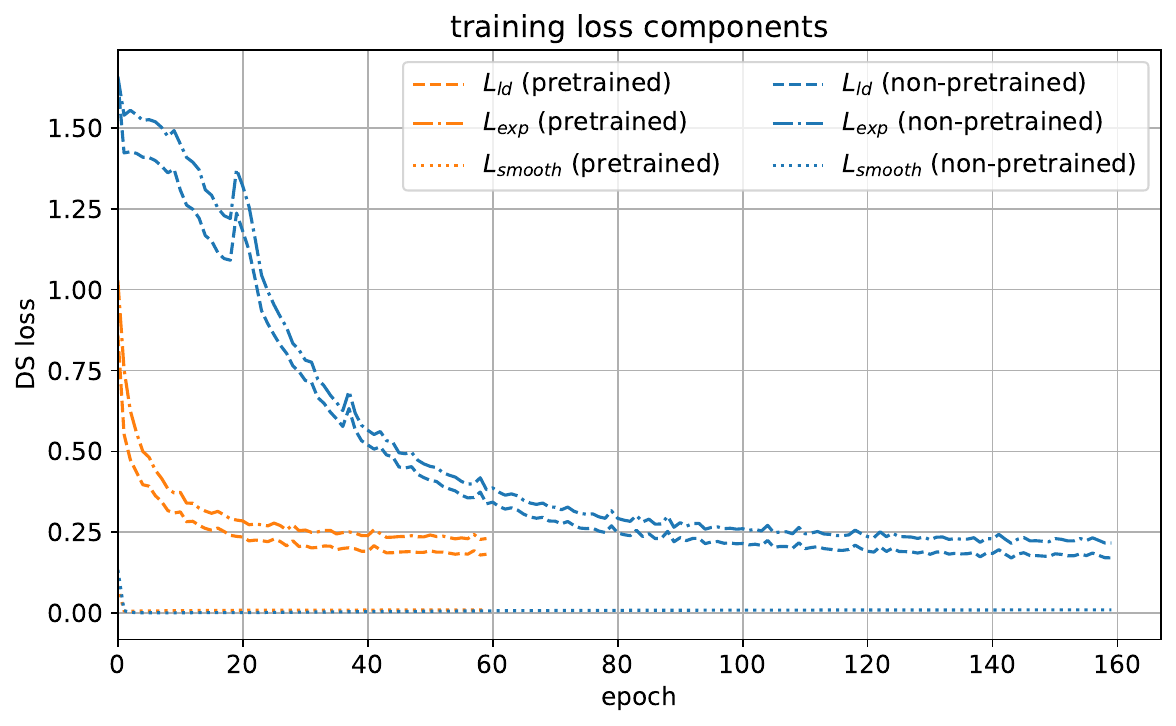}
            \subcaption{Training loss by epoch separated into the \num{3} loss terms (cf. Equation~\eqref{eq:loss_comp}). The blue lines show the loss components by eopch for the non-pretrained model, orange lines show the ones of the pretrained \sugarvit{} model.}
            \label{fig:sugarvit_pretrained_vs_non-pretrained:train_loss}
	\end{subfigure}
        \hfill
	\begin{subfigure}{\plotwidth}
		\includegraphics[width=\plotwidth]{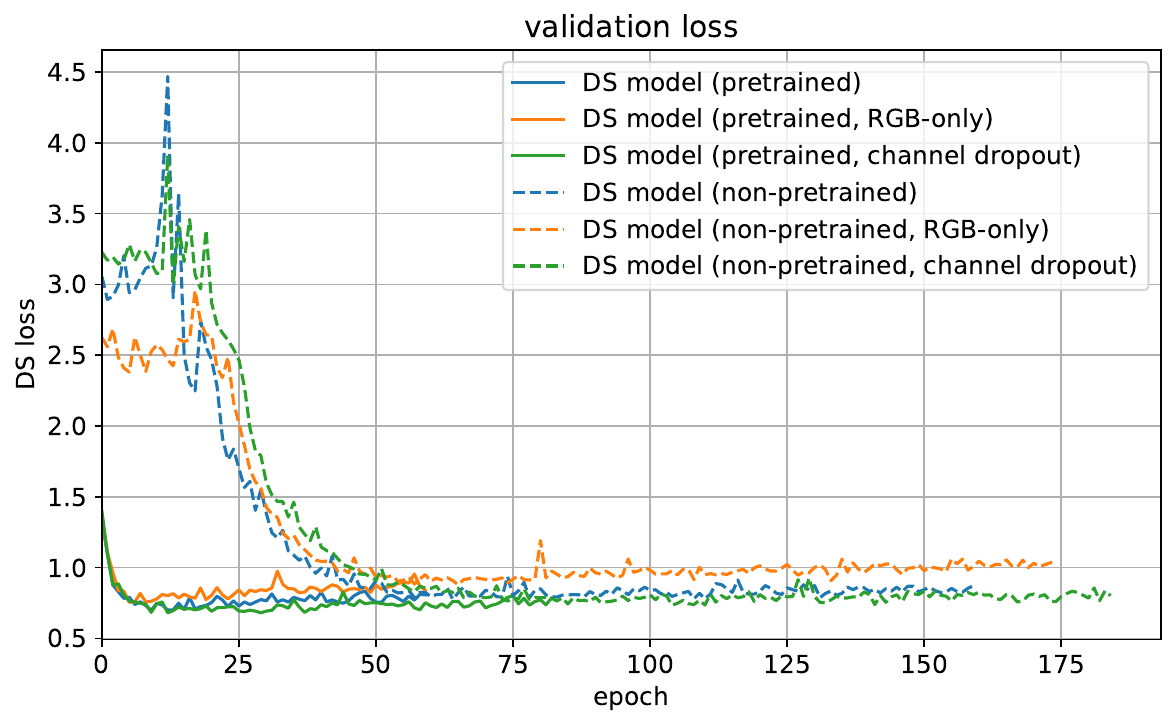}
            \subcaption{Validation loss by epoch for the pretrained and non-pretrained \sugarvit{} including the two training variants with channel dropout and \ac{RGB}-only information.}
            \label{fig:sugarvit_pretrained_vs_non-pretrained:val_loss}
	\end{subfigure}
        \hfill
	\begin{subfigure}{\plotwidth}
		\includegraphics[width=\plotwidth]{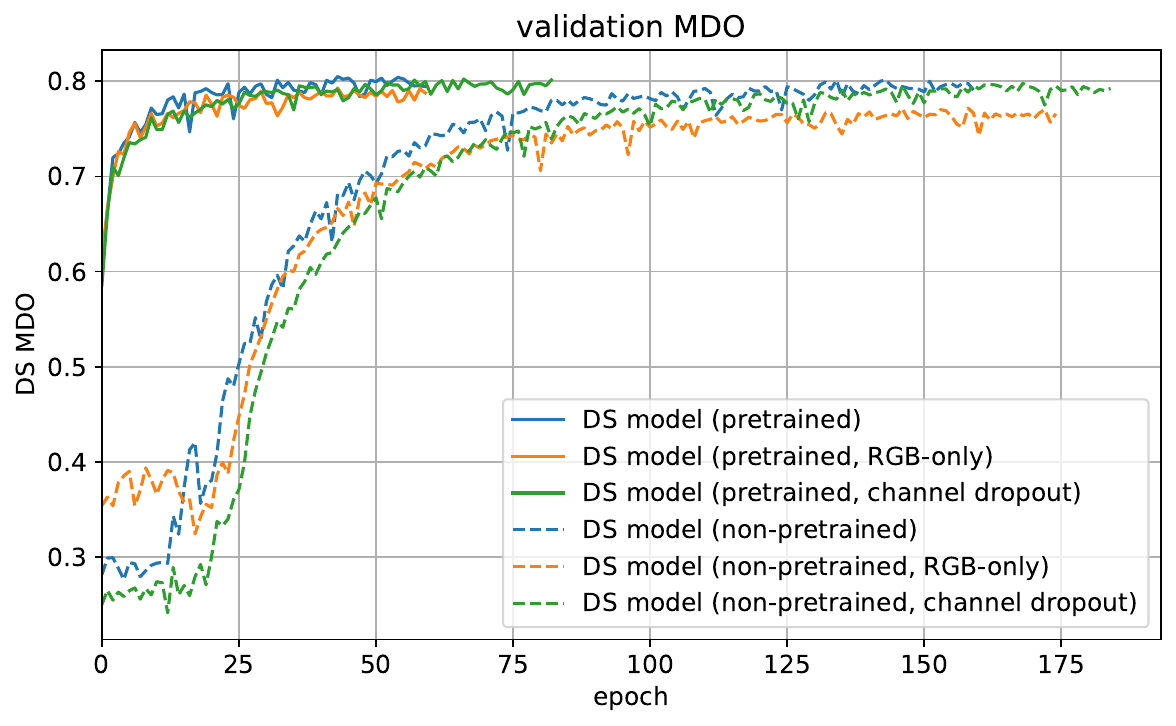}
            \subcaption{Validation \acf{MDO} by epoch for the pretrained and non-pretrained \sugarvit{} including the two training variants with channel dropout and \ac{RGB}-only information.}
            \label{fig:sugarvit_pretrained_vs_non-pretrained:val_mdo}
	\end{subfigure}
	\caption{Comparison of the pretrained and non-pretrained \sugarvit{} model.}
	\label{fig:sugarvit_pretrained_vs_non-pretrained}
\end{figure}

In this section, we want to discuss the training and validation metrics of the pretrained \sugarvit{} compared to a non-pretrained model that is trained \enquote{from scratch}. Also, we show results for the two training variants mentioned in Section~\ref{sec:results} by using channel dropout or only \ac{RGB} information during training. For channel dropout, the validation is done without dropout. Some performance plots are shown in Figure~\ref{fig:sugarvit_pretrained_vs_non-pretrained}. During the training process (cf. Figure~\ref{fig:sugarvit_pretrained_vs_non-pretrained:train_loss}), all training loss components converge to similar values, whereby the pretrained models converge, as expected, substantially faster. The plot only shows the full model. The loss curves for the other training variants are similar and can be found in the supplementary material~\ref{sec:appendix}. The validation loss (cf. Figure~\ref{fig:sugarvit_pretrained_vs_non-pretrained:val_loss}) shows similar behavior. In addition, a slight overfitting can be observed for all trainings as the loss is increasing after a minimum reach at about epochs \numrange{10}{15} for the pretrained and about epochs \numrange{50}{80} for the non-pretrained models. The overfitting can not be observed in terms of validation \ac{MDO} (cf. Figure~\ref{fig:sugarvit_pretrained_vs_non-pretrained:val_mdo}). For both validation loss and validation \ac{MDO} the pretrained models reach, besides the faster convergence, slightly better values compared to the non-pretrained models. Overall, the convergence for the channel dropout training is tendentially slower than for the \ac{RGB}-only and even slower compared to the full model.

So far, we just evaluated the model on the validation dataset which is, being a random subset of the training data, quite similar to the training data. In a next step, we want to evaluate our \sugarvit{} models on test data, that is completely unknown by the model in order to see the generalization capabilities of our approach.

\subsection{Evaluation on Test Dataset}~\label{sec:test_eval}

Conclusively, we want to evaluate our model on unseen data. Therefore, we use our dataset Te01 (cf.~\ref{tab:data_overview}). Although we stated that the \ac{MAE} is not an appropriate metric for the \ac{DLDL} approach in \sugarvit{}, it can give some insights on the prediction quality in the field usage, since the expectation value of the predicted label distribution is used as the overall model prediction. Recap, that the \ac{MAE} of $N$ \ac{DS} predictions is given by
\begin{align}
    \ac{MAE}(\ac{DS}) &= \frac{1}{N}\sum\limits_{i=1}^N\left|\ac{DS}_\text{true}(\bx_i) - \ac{DS}_\text{pred}(\bx_i)\right|\\
    &= \frac{1}{N}\sum\limits_{i=1}^N\left|E\left[\mP(y\mid\bx_i)\right] - E\left[\hat{\mP}(y\mid\bx_i)\right]\right|\,.
\end{align}
We evaluated three of our model adaptions, i.e., with using all information, with using channel dropout during training, and with only using \ac{RGB} channels, each of them with and without using pretraining. Certainly, recognitions can still be incorrect. However, we can apply some techniques to reduce the errors. On the one hand, we can augment the input and use the average of all augmented inputs as the final prediction for the augmented image. One could use any augmentation that we also used during training. However, we just use \enquote{simple} augmentations here like mirroring and rotation in order not to reduce the performance, i.e., inference time too much. Thus, we can augment one single image to \num{8} instances in total. Since the model outputs are \acp{pmf}, we can just add them and renormalize them by dividing by \num{8}. As stated in Section~\ref{sec:comparison_exp}, we could observe overfitting in the validation loss, while the \ac{MDO} is still increasing. We therefore evaluate our training variants by their best models concerning both metrics. It can be observed that the models with the best validation loss, respectively, perform slightly better in test data evaluation. Table~\ref{tab:test_mdo_mae:best_loss} shows the \ac{MAE} and \ac{MDO} values distributed in each true \ac{DS} label for those models. The values for the best validation \ac{MDO} models can be found in the supplementary material~\ref{sec:appendix}. In all evaluations, we use the augmentations mentioned above.

The most remarkable observation is that all models show performance shortcomings for plants with \ac{DS} of about \numrange{5}{9}. A reason might be the limited amount of training data in comparison to \acp{DS} that occur far more frequent than low and high \acp{DS}. We already try to cope with this imbalance by a weighted sampling as mentioned above. However, the augmented data surely is no actual \enquote{new} data in the actual sense. Unfortunately, those intermediate \ac{DS} scores are also the most ambiguous ones due to possible bias by margin of interpretation and evaluation. The data lack could be seen when comparing the results for the two 5-channel-trained models with the \ac{RGB}-only model. For \ac{DS} with many training data instances, the 5-channel models typically outperform the \ac{RGB} model, whereas for low-data regimes, the \ac{RGB} model outperforms the 5-channel models. This might be due to the lower complexity for \num{3} rather than \num{5} channels from that the features have to be extracted. Another remarkable result is that, apparently, the model channel-dropout does not show significant improvement in terms of generalization to unseen data. It outperforms the fully trained model in only a few \acp{DS}. In difference to the evaluation on validation data, the non-pretrained models outperform the pretrained ones in most cases for the test dataset. This is quite remarkable and opens the discussion if the pretraining possibly leads to a slight overfitting to the data and gives another hint about the importance of generalization, especially in agricultural-related use cases. Nevertheless, the idea of pretraining is still important, since training times can be reduced by adaption of one trained backbone to multiple labels of interest. Additionally, in use cases with low data coverage, finetuning existing, pretrained models might be the only way to get performant prediction models.

For the usage of \sugarvit{} in the field, there are still some points to mention regarding further prediction improvements that we want to discuss in the next section. 

\begin{table*}[t]
    \begin{center}
    \scriptsize
    \begin{minipage}{\columnwidth}
        \begin{tabular*}{\columnwidth}{l@{\extracolsep{\fill}}c@{\extracolsep{\fill}}c@{\extracolsep{\fill}}c@{\extracolsep{\fill}}c@{\extracolsep{\fill}}c@{\extracolsep{\fill}}c@{\extracolsep{\fill}}}\toprule
            \textbf{true DS} & \multicolumn{3}{c}{\textbf{non-pretrained}} & \multicolumn{2}{c}{\textbf{pretrained}} \\
            & full & ch. dropout & RGB & full & ch. dropout & RGB \\\midrule
            0             & \num{0.39}	& \textbf{\num{0.34}}	& \num{0.73}	& \textbf{\num{0.43}} & \num{0.50} &	\num{0.55} \\
            1             & \textbf{\num{0.50}}	& \num{0.56}	& \num{0.76}	& \textbf{\num{0.49}} & \num{0.57} &	\num{0.63} \\
            2             & \textbf{\num{0.82}}	& \num{0.91}	& \num{1.25}	& \textbf{\num{0.97}} & \num{1.09} &	\num{1.12} \\
            3             & \num{1.16}	& \textbf{\num{1.15}}	& \num{1.28}	& \num{1.26} & \textbf{\num{1.07}} &	\num{1.14} \\
            4             & \num{1.59}	& \textbf{\num{1.48}}	& \num{1.62}	& \num{1.67} & \textbf{\num{1.37}} &	\num{1.59} \\
            5             & \num{1.61}	& \textbf{\num{1.47}}	& \num{1.61}	& \num{1.56} & \textbf{\num{1.33}} &	\num{1.54} \\
            6             & \num{2.20}	& \num{2.20}	& \textbf{\num{1.93}}	& \num{2.08} & \textbf{\num{2.01}} &	\num{2.19} \\
            7             & \num{2.04}	& \num{2.04}	& \textbf{\num{1.72}}	& \num{1.99} & \num{1.59} &	\textbf{\num{1.55}} \\
            8             & \num{3.27}	& \num{3.11}	& \textbf{\num{3.03}}	& \num{3.02} & \textbf{\num{2.86}} &	\num{2.91} \\
            9             & \num{2.53}	& \num{2.46}	& \textbf{\num{2.25}}	& \num{2.35} & \num{2.24} &	\textbf{\num{2.18}} \\
            10            & \num{1.59}	& \num{1.53}	& \textbf{\num{1.43}}	& \num{1.45} & \textbf{\num{1.21}} &	\num{1.32} \\\midrule
            total         & \textbf{\num{0.67}}	& \num{0.70}	& \num{0.99}	& \textbf{\num{0.73}} & \num{0.81} &	\num{0.85} \\
            total (corr.) & \num{1.61}	& \textbf{\num{1.57}}	& \num{1.60}	& \num{1.57} & \textbf{\num{1.44}} &	\num{1.52} \\
            \bottomrule
        \end{tabular*}   
        \subcaption{\Acf{MAE}.}
    \end{minipage}
    \hfill
    \begin{minipage}{\columnwidth}
        \begin{tabular*}{\columnwidth}{l@{\extracolsep{\fill}}c@{\extracolsep{\fill}}c@{\extracolsep{\fill}}c@{\extracolsep{\fill}}c@{\extracolsep{\fill}}c@{\extracolsep{\fill}}c@{\extracolsep{\fill}}}\toprule
            \textbf{true DS} & \multicolumn{3}{c}{\textbf{non-pretrained}} & \multicolumn{2}{c}{\textbf{pretrained}} \\
            & full & ch. dropout & RGB & full & ch. dropout & RGB \\\midrule
            0             & \num{83.91} & \textbf{\num{88.62}} & \num{78.09} & \textbf{\num{82.25}} & \num{77.75} &	\num{80.65} \\
            1             & \textbf{\num{75.30}} & \num{73.52} & \num{66.42} & \textbf{\num{74.36}} & \num{73.37} &	\num{69.40} \\
            2             & \textbf{\num{55.95}} & \num{53.63} & \num{42.97} & \textbf{\num{51.00}} & \num{46.83} &	\num{46.69} \\
            3             & \textbf{\num{48.97}} & \num{48.34} & \num{45.11} & \num{46.83} & \textbf{\num{52.44}} &	\num{49.05} \\
            4             & \num{37.23} & \textbf{\num{38.84}} & \num{38.59} & \num{38.07} & \textbf{\num{44.10}} &	\num{38.12} \\
            5             & \num{34.43} & \num{37.87} & \textbf{\num{38.43}} & \num{34.29} & \textbf{\num{38.77}} &	\num{35.58} \\
            6             & \num{25.16} & \num{25.32} & \textbf{\num{28.82}} & \textbf{\num{23.83}} & \num{23.76} &	\num{21.98} \\
            7             & \num{28.20} & \num{26.46} & \textbf{\num{30.16}} & \num{26.88} & \num{32.69} &	\textbf{\num{32.89}} \\
            8             & \num{16.84} & \num{17.29} & \textbf{\num{17.89}} & \num{18.56} & \textbf{\num{22.72}} &	\num{21.20} \\
            9             & \num{31.25} & \num{33.01} & \textbf{\num{33.46}} & \num{35.67} & \num{36.35} &	\textbf{\num{36.88}} \\
            10            & \num{49.50} & \num{52.69} & \textbf{\num{53.93}} & \num{51.90} & \num{56.62} &	\textbf{\num{57.98}} \\\midrule
            total         & \num{68.10} & \textbf{\num{68.18}} & \num{59.16} & \textbf{\num{65.47}} & \num{62.70} &	\num{62.25} \\
            total (corr.) & \num{44.25} & \textbf{\num{45.05}} & \num{43.08} & \num{43.97} & \textbf{\num{45.95}} &	\num{44.58} \\
            \bottomrule
        \end{tabular*}   
        \subcaption{\Acf{MDO} in \si{\percent}.}
    \end{minipage}
    \caption{Evaluation metrics for test dataset separated by true \ac{DS} value. For each training variant, the model with the best validation loss is chosen. Below the single \acp{DS}, the metrics for the full dataset are listed, both with and without correcting for the label abundance.}
    \label{tab:test_mdo_mae:best_loss}
    \end{center}
\end{table*}

\subsection{Application in the Field}

Frameworks like~\cite{plantcataloging} help to extract the plant positions and extraction of the single images for large-scale \ac{UAV} image data. Thereafter, our trained \sugarvit{} model can be applied for new field experiments, enabling fast large-scale \ac{DS} annotations.

\begin{figure}[!htb]
	\centering
	\includegraphics[width=\plotwidth]{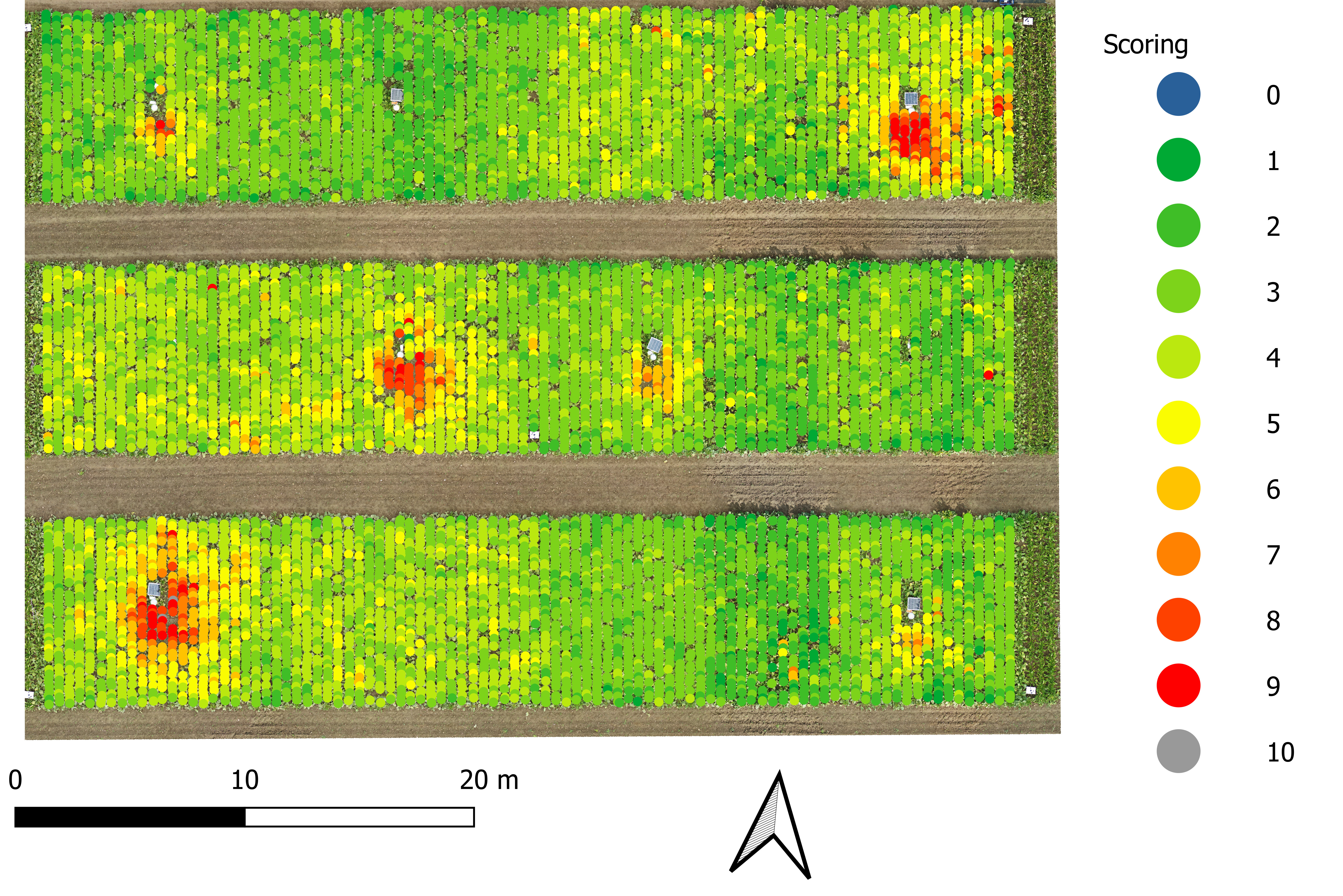}
	\caption{Exemplary application of \sugarvit{} for disease severity prediction on unseen \ac{UAV} data. Each prediction is completely independent of its surrounding predictions. The model shows a highly consistent prediction behavior.}
	\label{fig:test_qgis}
\end{figure}

 Figure~\ref{fig:test_qgis} shows an exemplary application of our model to a test dataset image using the described augmented evaluation. On the other hand, the model does not consider temporal and spatial dependency between the plant images so far. We could further reduce the error rate by correcting single \enquote{obvious} outliers that do not fit into the temporal and spatial vicinity of the other plants. Additionally, \sugarvit{} has the advantage to actually output a label distribution. Since we know, with which (fixed) training label standard deviation $\sigma_\text{train}$ the model is trained, we can compare the standard deviation of the output (assuming a normal distribution) $\sigma_\text{pred}$ with it in order to see, how \enquote{confident} the model is in its prediction. Thus, for each output, we can calculate a \enquote{confidence}
\begin{align}
    \tilde{c} = \frac{\sigma_\text{pred}}{\sigma_\text{train}}\,,
\end{align}
that is $1$ for an exact conformity of standard deviations. For values $\tilde{c}>1$, the model is more and more unconfident, whereas for values $0<\tilde{c}<1$, the model is over-confident in its decision. However, please note that this is no proper confidence in a statistical sense, since the expectation value still could be completely wrong. Since we cannot know the expectation value in an unlabeled dataset, this purely standard deviation based value is a rather imprecise yet helpful measure of the model's confidence. 

In our model training and evaluation framework, we include some convenience functions to load orthoimages and plant positions as \textit{geopackage} or similar files and evaluate models on. Afterward, the prediction can be exported again to \textit{geopackage} format or, for instance, as \textit{Pandas DataFrame} objects. Thus, we added interfaces to widely used \textit{GIS} software that are frequently used for georeferenced image data.

\subsection{Attention Maps}

\begin{figure*}[!htb]
	\centering
	\includegraphics[width=\textwidth]{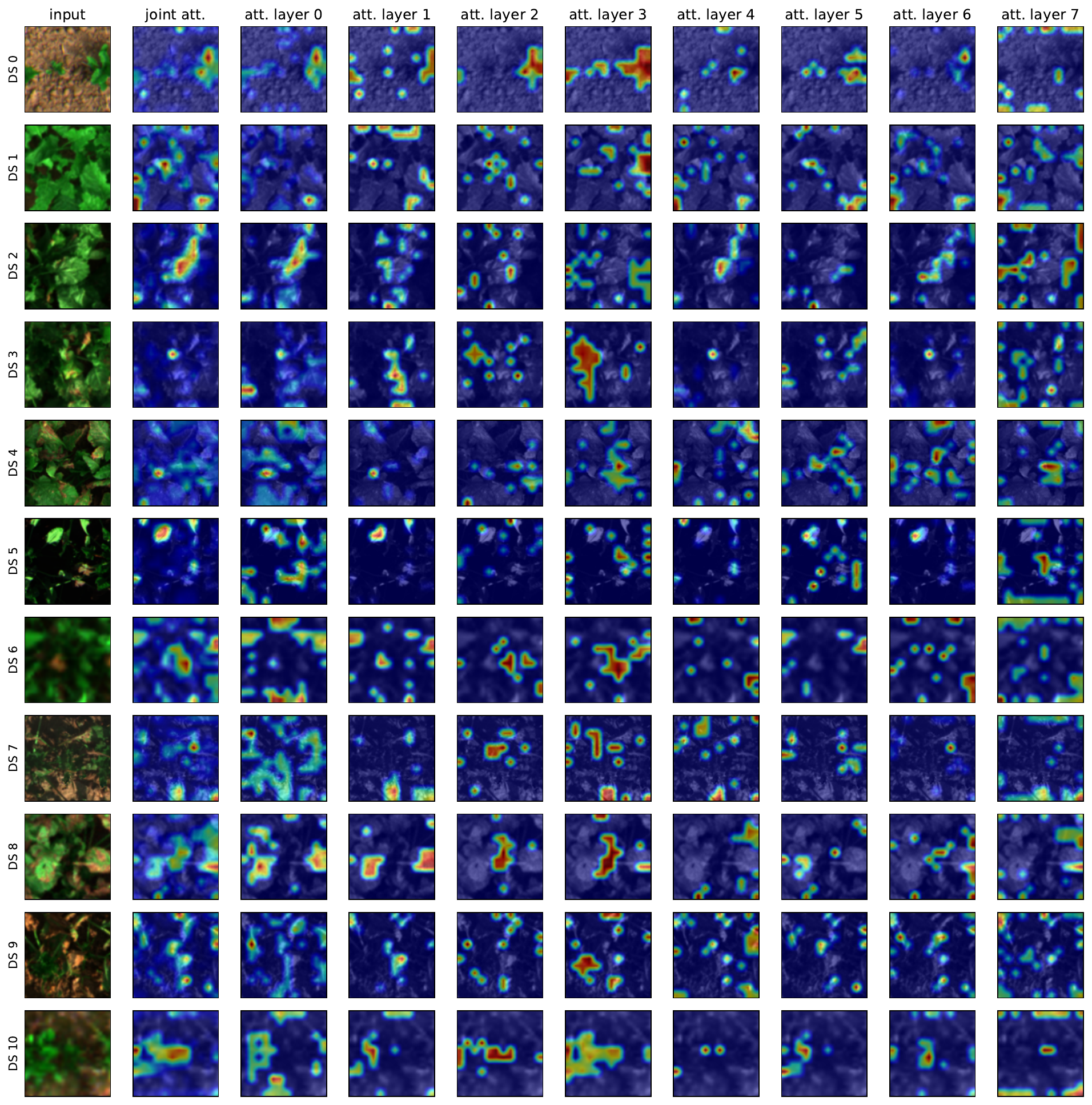}
	\caption{Attention maps for an example image per disease severity class. The first column shows the original input image in its \ac{RGB} representation. The second column is the joint attention map after performing attention rollout~\cite{attention_rollout}. The following columns are the attention maps of each of the \num{8} attention layers in our \sugarvit{} model.}
	\label{fig:attention_maps}
\end{figure*}

The fact, that the backbone of our \sugarvit{} model is based on the attention mechanism~\cite{attention}, we can analyze and, ideally, interpret which image feature are more or less important to the model's decisions. One helpful visualization method for that are so-called \textit{attention maps}~\cite{attention_rollout}. Roughly spoken, attention maps can visualize, \enquote{where the model looks at}. The \ac{ViT} backbone in \sugarvit{} consists of \num{8} attention layers, with \num{4} attention heads each, that can in principle be trained to focus on completely different features. In order to accumulate the attention maps of each single layer to one overall map, the technique of \textit{attention rollout}~\cite{attention_rollout} is used. Figure~\ref{fig:attention_maps} shows the result for one randomly chosen image from the validation dataset per \ac{DS} class. A main observation is that \sugarvit{} indeed focuses on the plant itself and not, e.g., on the amount of visible soil around it. This is particularly visible for the \ac{DS} \num{0} example. Additionally, one observes that the model focuses on multiple image regions that seem to be complementary for the decision-making process, which is exactly the power of the attention mechanism compared to \acp{CNN}. \acp{CNN} learn filters that are applied on the whole image. Thus, relations between pixel values can only be covered locally. The attention mechanism allows connecting those local features with other distant images regions and, thus, is considered to have more power of \enquote{understanding} the image as a whole.

\section{Discussion}

Our work shows, that simple normalizing methods like standardization can already outperform more sophisticated and expensive normalization methods like histogram equalization (cf. Section~\ref{sec:norm_exp}). For our use-case where spectral differences carry much information, a total standardization leads to better results than a channel-wise standardization. However, it has to be stated, that the data used in this work has been calibrated by a reflectance panel, so the spectral information is directly comparable. Thus, if the data can not be calibrated by any reason, also the channel-wise standardization may be a good choice, since it also leads to acceptable prediction qualities.

The pretraining on environmental metadata turned out to be beneficial for the final prediction quality as well as the training speed (cf. Section~\ref{sec:sugarvit_exp}). Additionally, our pretraining on general, environmental annotation and the subsequent finetuning on the annotation of interest can be an approach for further generalization even on smaller datasets. The pretrained \ac{ViT} backbone can be seen as a fixed, plant-image-aware feature extractor that learned plant specific traits. On top of that, a smaller model can be trained for different annotation purposes, which accelerates and improves the training procedure substantially. This concept is also widely used in use cases of large language models where the model sizes often exceed computational resources for local training. Pretraining also enables the usage of large-scale image data. Even if only few data is labeled with expensive human-expert annotations, unlabeled data can still be used in a pretraining stage unfolding the full potential of the collected data. If data labeling originated from multiple human experts that may have ambiguous estimations or assessment guidelines, our usage of \ac{LDL} is able to incorporate detailed uncertainty information in training process and, finally, in the model.

The use of attention mechanism instead of convolutional networks turns out to make sense in this use case because of the long-distance relations of leaf spots. The interpretability of resulting attention maps on single instances is often questionable. However, they can reveal, if the model focuses on the right regions and features itself rather than on spurious correlations.

\section{Conclusions}

Generalization to data from unseen fields with unseen weather conditions and climate is certainly one of the most challenging questions in data-driven machine learning approaches in agriculture. Our findings in Section~\ref{sec:test_eval} emphasize that. The data given in the scope of this work comprises already \num{4} growing seasons but only from some locations in Central Germany. If the model performs comparably well in other regions is at least questionable. Nevertheless, a model that is at least locally accurate already has a high value for increasing the efficiency of \ac{DS} assessment. Extrapolation to other environmental conditions is difficult, but interpolation on the same field has the potential to save valuable expert working time in the field, where the model can complement a few spot-wise expert annotations on the whole field. As seen in Section~\ref{sec:test_eval}, the high data imbalance and label ambiguity still remains challenging, even with our contributions of weighed sampling and \ac{LDL}, respectively. The disease assessment in individual objects as plants is hard to standardize and schematize by exemplary images, as we, for instance, have in the case of~\ac{CLS}. Therefore, it is important to have a model that incorporates label uncertainties and is transparent in its prediction uncertainties, like in \sugarvit{}.

With this large-scale \ac{DS} assessment available, further challenges regarding disease assessment can be tackled. The retrieval of \ac{DS} complemented by further environmental sensors enables, for instance, detailed investigations on disease spread and its modeling. Consequently, our approach could also find application in terms of disease control by, for instance, punctual application of pesticides, lowering costs and environmental impact. Thus, future work regarding this topic will be to use \sugarvit{} for disease spread modeling. Perspectively, the concept behind \sugarvit{} could also be applied in a wide variety of other use cases in the field of UAV-supported phenotyping.~\cite{facu_rhizoctonia,multispectral_phenotyping_1,uav_prec_agri_survey}

\section{Abbreviations}

\begin{acronym}
    \acro{CLS}{Cercospora Leaf Spot}
    \acro{CNN}{Convolutional Neural Network}
    \acro{DAC}{days after canopy closure}
    \acro{DLDL}{Deep Label Distribution Learning}
    \acro{DS}{disease severity}
    \acro{FCL}{fully-connected layer}
    \acro{GDD}{Growing Degree Day}
    \acro{GPT}{Generative Pretrained Transformer}
    \acro{HE}{Histogram Equalization}
    \acro{KLD}{Kullback-Leibler Divergence}
    \acro{LDL}{Label Distribution Learning}
    \acro{MAE}{mean absolute error}
    \acro{MDO}{mean distribution overlap}
    \acro{MLP}{Multi-Layer Perceptron}
    \acro{NLP}{natural language processing}
    \acro{NPG}{number of possible generation}
    \acro{pmf}{probability mass function}
    \acro{RGB}{red-, green-, and blue-channel}
    \acro{std}{standard deviation}
    \acro{UAV}{Unmanned Aerial Vehicle}
    \acro{ViT}{Vision Transformer}
\end{acronym}

\section{Competing Interests}
The authors declare that they have no competing interests.

\section{Funding}

This work has been funded by the Deutsche Forschungsgemeinschaft (DFG, German Research Foundation) under Germany’s Excellence Strategy – EXC 2070 – 390732324.

\section{Authors' Contributions}

MG developed the theoretical formalism and the model, preprocessed the data for model training and evaluation, performed the experiments in this work, and wrote most of the manuscript. FRIY and AABA contributed in the sections regarding use case and remote sensing, prepared and provided the image and environmental data used for the model training and evaluation. AKM, RS, and CB supervised the project and advised on the manuscript preparation.



\bibliographystyle{plain}  
\bibliography{references}  

\clearpage
\appendix
\input{supplementary}

\end{document}

%% file: supplementary.tex
\section{Appendices}\label{sec:appendix}

\subsection{Additional Plots for \sugarvit{} Pretraining}

\begin{figure}[!htb]
	\centering
	\begin{subfigure}{\plotwidth}
		\includegraphics[width=\plotwidth]{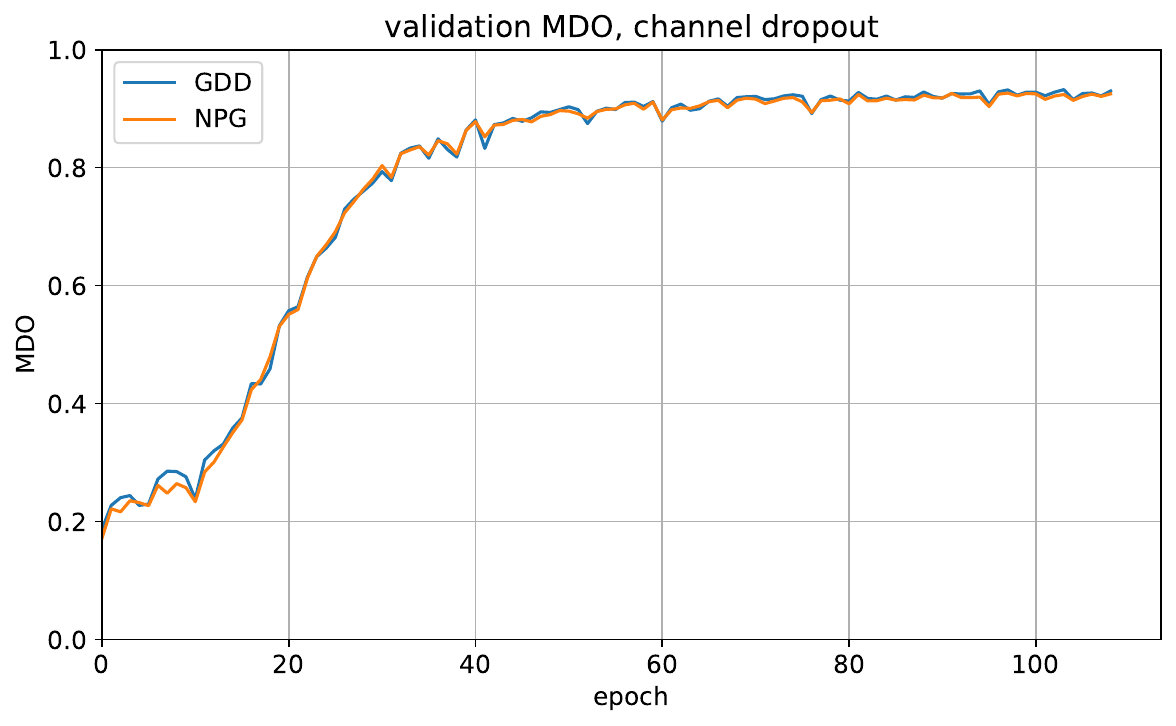}
            \subcaption{Validation \acf{MDO} by epoch for both \ac{GDD} and \ac{NPG} labels with channel dropout during training.}
	\end{subfigure}
        \hfill
	\begin{subfigure}{\plotwidth}
		\includegraphics[width=\plotwidth]{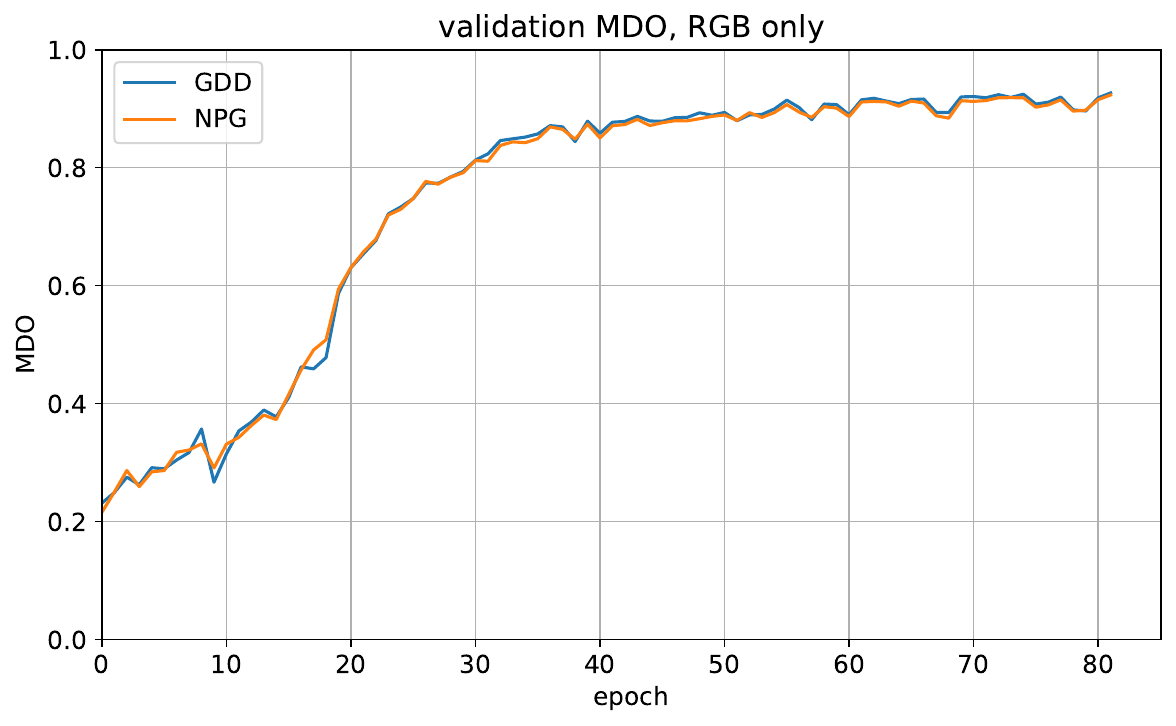}
            \subcaption{Validation \acf{MDO} by epoch for both \ac{GDD} and \ac{NPG} labels with only \ac{RGB} channel information.}
	\end{subfigure}
	\caption{Results of the \sugarvit{} pretraining for the channel dropout and the \ac{RGB}-only variants.}
\end{figure}

\subsection{Additional Plots for \ac{DS} Training of \sugarvit{}}

\begin{figure}[!htb]
	\centering
	\begin{subfigure}{\plotwidth}
		\includegraphics[width=\plotwidth]{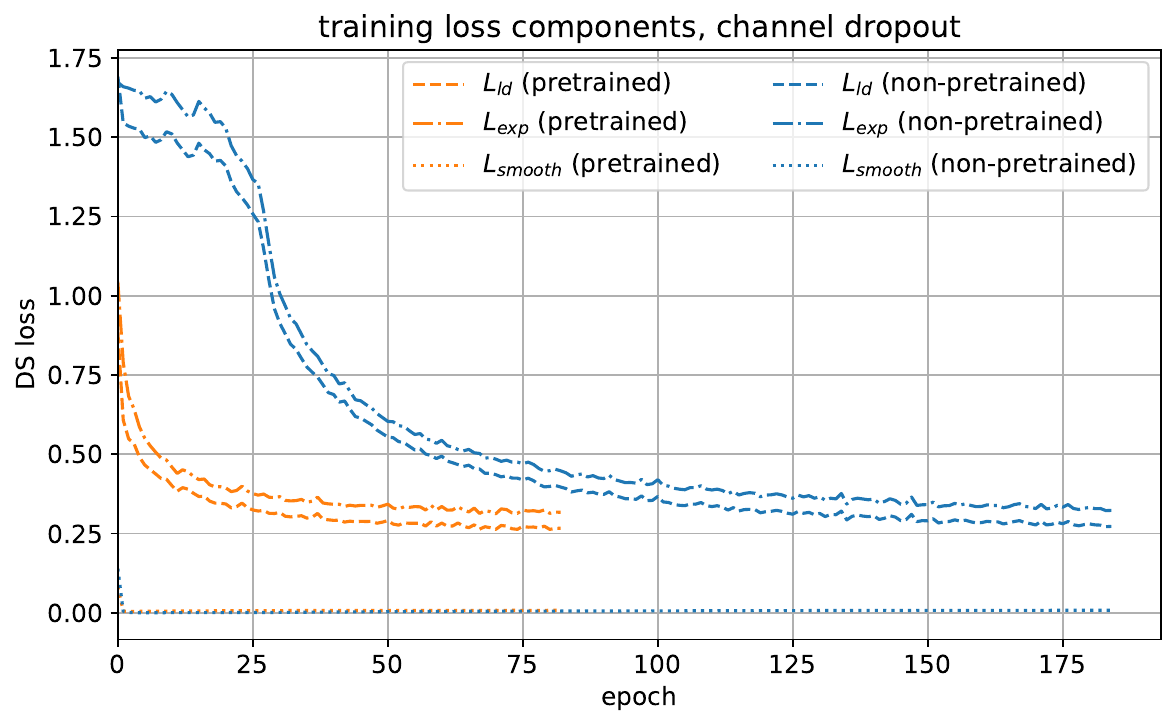}
            \subcaption{Training loss components by training epoch with channel dropout.}
	\end{subfigure}
        \hfill
	\begin{subfigure}{\plotwidth}
		\includegraphics[width=\plotwidth]{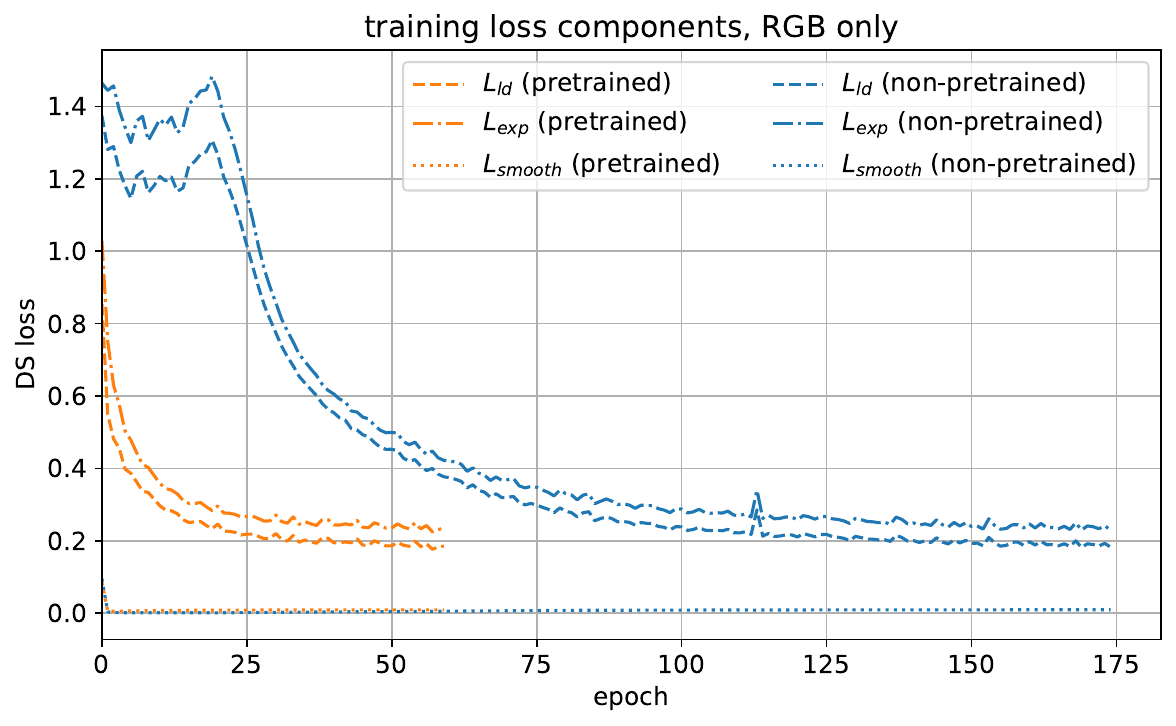}
            \subcaption{Training loss components by training epoch with only \ac{RGB} channel information.}
	\end{subfigure}
	\caption{Results of the \sugarvit{} \ac{DS} training for the channel dropout and the \ac{RGB}-only variants.}
\end{figure}

\subsection{Additional Test Evaluation Metrics}

\begin{table*}[t]
    \begin{center}
    \scriptsize
    \begin{minipage}{\columnwidth}
        \begin{tabular*}{\columnwidth}{l@{\extracolsep{\fill}}c@{\extracolsep{\fill}}c@{\extracolsep{\fill}}c@{\extracolsep{\fill}}c@{\extracolsep{\fill}}c@{\extracolsep{\fill}}c@{\extracolsep{\fill}}}\toprule
            \textbf{true DS} & \multicolumn{3}{c}{\textbf{non-pretrained}} & \multicolumn{2}{c}{\textbf{pretrained}} \\
            & full & ch. dropout & RGB & full & ch. dropout & RGB \\\midrule
            0  & \textbf{\num{0.38}} & \num{0.45} & \num{0.60} & \textbf{\num{0.57}} & \num{0.59} & \num{0.69} \\
            1  & \num{0.46} & \textbf{\num{0.44}} & \num{0.70} & \num{0.54} & \textbf{\num{0.50}} & \num{0.51} \\
            2  & \textbf{\num{0.76}} & \num{0.77} & \num{1.26} & \textbf{\num{0.87}} & \textbf{\num{0.87}} & \num{1.08} \\
            3  & \textbf{\num{0.97}} & \num{1.02} & \num{1.47} & \textbf{\num{1.16}} & \num{1.23} & \num{1.38} \\
            4  & \textbf{\num{1.45}} & \num{1.46} & \num{1.86} & \num{1.73} & \textbf{\num{1.69}} & \num{1.74} \\
            5  & \num{1.74} & \textbf{\num{1.63}} & \num{1.90} & \num{1.69} & \textbf{\num{1.67}} & \num{1.72} \\
            6  & \num{2.37} & \num{2.39} & \textbf{\num{2.30}} & \num{2.26} & \textbf{\num{2.08}} & \num{2.29} \\
            7  & \num{2.40} & \num{2.02} & \textbf{\num{1.76}} & \num{2.09} & \num{2.21} & \textbf{\num{1.57}} \\
            8  & \num{3.52} & \num{3.34} & \textbf{\num{2.91}} & \num{3.19} & \num{2.99} & \textbf{\num{2.96}} \\
            9  & \num{2.80} & \num{2.45} & \textbf{\num{2.00}} & \num{2.63} & \num{2.36} & \textbf{\num{2.20}} \\
            10 & \num{1.61} & \num{1.40} & \textbf{\num{1.32}} & \textbf{\num{1.44}} & \num{1.50} & \num{1.56} \\\midrule
            total & \textbf{\num{0.63}} & \num{0.65} & \num{0.96} & \textbf{\num{0.75}} & \textbf{\num{0.75}} & \num{0.86} \\
            total (corr.) & \num{1.68} & \textbf{\num{1.59}} & \num{1.64} & \num{1.65} & \textbf{\num{1.61}} & \textbf{\num{1.61}} \\
            \bottomrule
        \end{tabular*}   
        \subcaption{\Acf{MAE}.}
    \end{minipage}
    \hfill
    \begin{minipage}{\columnwidth}
        \begin{tabular*}{\columnwidth}{l@{\extracolsep{\fill}}c@{\extracolsep{\fill}}c@{\extracolsep{\fill}}c@{\extracolsep{\fill}}c@{\extracolsep{\fill}}c@{\extracolsep{\fill}}c@{\extracolsep{\fill}}}\toprule
            \textbf{true DS} & \multicolumn{3}{c}{\textbf{non-pretrained}} & \multicolumn{2}{c}{\textbf{pretrained}} \\
            & full & ch. dropout & RGB & full & ch. dropout & RGB \\\midrule
            0  & \textbf{\num{82.92}} & \num{81.40} & \num{78.71} & \num{76.46} & \textbf{\num{76.99}} & \num{76.30} \\
            1  & \textbf{\num{76.49}} & \num{76.36} & \num{68.63} & \num{72.51} & \num{71.48} & \textbf{\num{74.81}} \\
            2  & \textbf{\num{59.30}} & \num{58.14} & \num{45.05} & \num{55.43} & \textbf{\num{56.05}} & \num{50.34} \\
            3  & \textbf{\num{54.91}} & \num{52.29} & \num{44.52} & \textbf{\num{50.30}} & \num{46.94} & \num{44.93} \\
            4  & \textbf{\num{40.75}} & \num{40.58} & \num{35.33} & \num{33.40} & \textbf{\num{33.68}} & \num{33.45} \\
            5  & \num{30.13} & \textbf{\num{34.08}} & \num{31.21} & \num{30.50} & \num{32.94} & \textbf{\num{33.08}} \\
            6  & \num{18.64} & \num{18.29} & \textbf{\num{20.17}} & \num{22.45} & \textbf{\num{24.21}} & \num{22.06} \\
            7  & \num{20.34} & \num{27.52} & \textbf{\num{30.44}} & \num{25.94} & \num{27.54} & \textbf{\num{34.40}} \\
            8  & \num{14.33} & \num{15.44} & \textbf{\num{21.26}} & \num{17.35} & \num{18.59} & \textbf{\num{19.02}} \\
            9  & \num{29.76} & \num{31.27} & \textbf{\num{38.83}} & \num{31.06} & \num{33.05} & \textbf{\num{34.34}} \\
            10 & \num{47.49} & \textbf{\num{55.64}} & \num{53.87} & \textbf{\num{52.95}} & \num{51.51} & \num{50.70} \\\midrule
            total & \textbf{\num{69.57}} & \num{68.60} & \num{60.63} & \num{64.98} & \textbf{\num{65.00}} & \num{63.47} \\
            total (corr.) & \num{43.19} & \textbf{\num{44.64}} & \num{42.55} & \num{42.58} & \num{43.00} & \textbf{\num{43.04}} \\
            \bottomrule
        \end{tabular*}   
        \subcaption{\Acf{MDO} in \si{\percent}.}
    \end{minipage}
    \caption{Evaluation metrics for test dataset separated by true \ac{DS} value. For each training variant, the model with the best validation \acf{MDO} is chosen. Below the single \acp{DS}, the metrics for the full dataset are listed, both with and without correcting for the label abundance.}
    \label{tab:test_mdo_mae:best_mdo}
    \end{center}
\end{table*}

%% file: main.bbl
\begin{thebibliography}{10}

\bibitem{npg_formula}
Sugar beet disease models.
\newblock \url{https://metos.at/en/disease-models-sugar-beet/}.
\newblock accessed: 07/31/2023.

\bibitem{attention_rollout}
Samira Abnar and Willem Zuidema.
\newblock Quantifying attention flow in transformers, 2020.

\bibitem{clahe_tomography}
Zohair Al-Ameen, Ghazali Sulong, Amjad Rehman, Abdullah Al-Dhelaan, Tanzila
  Saba, and Mznah Al-Rodhaan.
\newblock An innovative technique for contrast enhancement of computed
  tomography images using normalized gamma-corrected contrast-limited adaptive
  histogram equalization.
\newblock {\em {EURASIP} Journal on Advances in Signal Processing}, 2015(1),
  April 2015.

\bibitem{uav_prec_agri_survey}
Muhammet~Fatih Aslan, Akif Durdu, Kadir Sabanci, Ewa Ropelewska, and
  Seyfettin~Sinan Gültekin.
\newblock A comprehensive survey of the recent studies with uav for precision
  agriculture in open fields and greenhouses.
\newblock {\em Applied Sciences}, 12(3):1047, Jan 2022.

\bibitem{abel_disease_incidence}
Abel Barreto, Facundo~Ram\'{o}n Ispizua~Yamati, Mark Varrelmann, Stefan Paulus,
  and Anne-Katrin Mahlein.
\newblock Disease incidence and severity of cercospora leaf spot in sugar beet
  assessed by multispectral unmanned aerial images and machine learning.
\newblock {\em Plant Disease}, 107(1):188--200, 2023.
\newblock PMID: 35581914.

\bibitem{bleiholder_1}
H.~Bleiholder and H.~C. Weltzien.
\newblock Beiträge zur epidemiologie von cercospora beticola, sacc. an
  zuckerrübe - i. die inkubations- und die fruktifikationszeit.
\newblock {\em Journal of Phytopathology}, 72(4):344--353, 1971.

\bibitem{bleiholder_2}
H.~Bleiholder and H.~C. Weltzien.
\newblock Beiträge zur epidemiologie von cercospora beticola sacc. an
  zuckerrübe - ii. die konidienbildung in abhängigkeit von den
  umweltbedingungcn temperatur, relative luftfeuchtigkeit und licht.
\newblock {\em Journal of Phytopathology}, 73(1):46--68, 1972.

\bibitem{phytopathometry}
Clive~H. Bock, Jayme G.~A. Barbedo, Anne-Katrin Mahlein, and Emerson M.~Del
  Ponte.
\newblock A special issue on phytopathometry {\textemdash} visual assessment,
  remote sensing, and artificial intelligence in the twenty-first century.
\newblock {\em Tropical Plant Pathology}, 47(1):1--4, February 2022.

\bibitem{multispectral_phenotyping_1}
Walter Chivasa, Onisimo Mutanga, and Chandrashekhar Biradar.
\newblock Uav-based multispectral phenotyping for disease resistance to
  accelerate crop improvement under changing climate conditions.
\newblock {\em Remote Sensing}, 12(15):2445, Jul 2020.

\bibitem{imagenet}
Jia Deng, Wei Dong, Richard Socher, Li-Jia Li, Kai Li, and Li~Fei-Fei.
\newblock Imagenet: A large-scale hierarchical image database.
\newblock In {\em 2009 IEEE conference on computer vision and pattern
  recognition}, pages 248--255. Ieee, 2009.

\bibitem{vit}
Alexey Dosovitskiy, Lucas Beyer, Alexander Kolesnikov, Dirk Weissenborn,
  Xiaohua Zhai, Thomas Unterthiner, Mostafa Dehghani, Matthias Minderer, Georg
  Heigold, Sylvain Gelly, Jakob Uszkoreit, and Neil Houlsby.
\newblock An image is worth 16x16 words: Transformers for image recognition at
  scale, 2020.

\bibitem{dldl}
Bin~Bin Gao, Chao Xing, Chen~Wei Xie, Jianxin Wu, and Xin Geng.
\newblock {Deep Label Distribution Learning with Label Ambiguity}.
\newblock {\em IEEE Transactions on Image Processing}, 26(6):2825--2838, 2017.

\bibitem{ldl}
Xin Geng.
\newblock Label distribution learning, 2016.

\bibitem{ldl_pose}
Xin Geng, Xin Qian, Zengwei Huo, and Yu~Zhang.
\newblock Head pose estimation based on multivariate label distribution.
\newblock {\em IEEE Transactions on Pattern Analysis and Machine Intelligence},
  44(4):1974--1991, 2022.

\bibitem{ldl_age}
Xin Geng, Chao Yin, and Zhi-Hua Zhou.
\newblock Facial age estimation by learning from label distributions.
\newblock {\em {IEEE} Transactions on Pattern Analysis and Machine
  Intelligence}, 35(10):2401--2412, oct 2013.

\bibitem{plantcataloging}
Maurice Günder, Facundo~R Ispizua Yamati, Jana Kierdorf, Ribana Roscher,
  Anne-Katrin Mahlein, and Christian Bauckhage.
\newblock {Agricultural plant cataloging and establishment of a data framework
  from UAV-based crop images by computer vision}.
\newblock {\em GigaScience}, 11:giac054, 06 2022.

\bibitem{full_kl_loss}
Maurice Günder, Nico Piatkowski, and Christian Bauckhage.
\newblock Full kullback-leibler-divergence loss for hyperparameter-free label
  distribution learning, 2022.

\bibitem{gdd_formula}
Carlyle~D. Holen and Alan~G. Dexter.
\newblock A growing degree day equation for early sugarbeet leaf stages.
\newblock {\em Sugarbeet Research and Extension Reports}, 27:152--157, 1997.

\bibitem{facu_sugarindustry}
Facundo~Ram\'{o}n Ispizua~Yamati, Abel Barreto, Maurice G{\"u}nder, Christian
  Bauckhage, and Anne-Katrin Mahlein.
\newblock Sensing the occurrence and dynamics of cercospora leaf spot disease
  using {UAV-supported} image data and deep learning.
\newblock {\em Zuckerindustrie}, pages 79--86, February 2022.

\bibitem{facu_rhizoctonia}
Facundo~Ram\'{o}n Ispizua~Yamati, Maurice G\"{u}nder, Abel~Andree
  Barreto~Alc\'{a}ntara, Jonas B\"{o}mer, Daniel Laufer, Christian Bauckhage,
  and Anne-Katrin Mahlein.
\newblock Automatic scoring of rhizoctonia crown and root rot affected sugar
  beet fields from orthorectified uav images using machine learning.
\newblock {\em Plant Disease}, 0(ja):null, 0.
\newblock PMID: 37755420.

\bibitem{kl_div}
James~M. Joyce.
\newblock {\em Kullback-Leibler Divergence}, pages 720--722.
\newblock Springer Berlin Heidelberg, Berlin, Heidelberg, 2011.

\bibitem{transformers_in_nlp}
Katikapalli~Subramanyam Kalyan, Ajit Rajasekharan, and Sivanesan Sangeetha.
\newblock Ammus : A survey of transformer-based pretrained models in natural
  language processing, 2021.

\bibitem{cercospora_kws}
Kleinwanzlebener Saatzucht AG, Rabbethge and Giesecke, Einbeck, Germany.
\newblock {\em Cercospora Tafel}, 1970.

\bibitem{transformers_survey}
Tianyang Lin, Yuxin Wang, Xiangyang Liu, and Xipeng Qiu.
\newblock A survey of transformers, 2021.

\bibitem{disease_assessment}
Forrest Nutter, Jr, Paul Teng, and F.M. Shokes.
\newblock Disease assessment terms and concepts.
\newblock {\em Plant Disease}, 75:1187--1188, 01 1991.

\bibitem{hist_eq_study}
Omprakash Patel, Yogendra P.~S. Maravi, and Sanjeev Sharma.
\newblock A comparative study of histogram equalization based image enhancement
  techniques for brightness preservation and contrast enhancement.
\newblock {\em CoRR}, abs/1311.4033, 2013.

\bibitem{ahe}
Stephen~M. Pizer, E.~Philip Amburn, John~D. Austin, Robert Cromartie, Ari
  Geselowitz, Trey Greer, Bart {ter Haar Romeny}, John~B. Zimmerman, and Karel
  Zuiderveld.
\newblock Adaptive histogram equalization and its variations.
\newblock {\em Computer Vision, Graphics, and Image Processing},
  39(3):355--368, 1987.

\bibitem{cls_2}
Lorena~I. Rangel, Rebecca~E. Spanner, Malaika~K. Ebert, Sarah~J. Pethybridge,
  Eva~H. Stukenbrock, Ronnie de~Jonge, Gary~A. Secor, and Melvin~D. Bolton.
\newblock Cercospora beticola: The intoxicating lifestyle of the leaf spot
  pathogen of sugar beet.
\newblock {\em Molecular Plant Pathology}, 21(8):1020--1041, 2020.

\bibitem{attention}
Ashish Vaswani, Noam Shazeer, Niki Parmar, Jakob Uszkoreit, Llion Jones,
  Aidan~N Gomez, \L~ukasz Kaiser, and Illia Polosukhin.
\newblock Attention is all you need.
\newblock In I.~Guyon, U.~Von Luxburg, S.~Bengio, H.~Wallach, R.~Fergus,
  S.~Vishwanathan, and R.~Garnett, editors, {\em Advances in Neural Information
  Processing Systems}, volume~30. Curran Associates, Inc., 2017.

\bibitem{ds_classification}
Guan Wang, Yu~Sun, and Jianxin Wang.
\newblock Automatic image-based plant disease severity estimation using deep
  learning.
\newblock {\em Computational Intelligence and Neuroscience}, 2017:1--8, 07
  2017.

\bibitem{cercospora}
John Weiland and Georg Koch.
\newblock Sugarbeet leaf spot disease (cercospora beticola sacc.).
\newblock {\em Molecular plant pathology}, 5(3):157--166, 2004.

\bibitem{cls_1}
John Weiland and Georg Koch.
\newblock Sugarbeet leaf spot disease (cercospora beticola sacc.)†.
\newblock {\em Molecular Plant Pathology}, 5(3):157--166, 2004.

\bibitem{fungal_leaf_disease_management}
P.~F.~J. Wolf and J.~A. Verreet.
\newblock An integrated pest management system in germany for the control of
  fungal leaf diseases in sugar beet: The {IPM} sugar beet model.
\newblock {\em Plant Disease}, 86(4):336--344, April 2002.

\bibitem{multispectral_phenotyping_2}
Rui Xu, Changying Li, and Andrew~H. Paterson.
\newblock Multispectral imaging and unmanned aerial systems for cotton plant
  phenotyping.
\newblock {\em PLOS ONE}, 14(2):1--20, 02 2019.

\end{thebibliography}
